\documentclass[twoside,11pt]{article}
\usepackage{jair}
\usepackage{theapa}
\usepackage{amsmath}
\usepackage{color}
\usepackage[normalem]{ulem}
\usepackage{amssymb}            
\usepackage{enumitem}           
\usepackage{graphicx}           
\usepackage{pbox}               
\usepackage{tikz}               
\makeatletter
\renewcommand{\jairheading}[5]{\def\ps@jairtps{\let\@mkboth\@gobbletwo%
\def\@oddhead{\scriptsize Journal of Artificial Intelligence Research #1 (#2) #3 \hfill Submitted #4; published #5}%
\def\@oddfoot{\scriptsize \copyright #2 National Research Council Canada. Reprinted with permission. \hfill}%
\def\@evenhead{}\def\@evenfoot{}}%
\thispagestyle{jairtps}}
\makeatother
\jairheading{44}{2012}{533-585}{03/12}{07/12}
\ShortHeadings{Domain and Function: A Dual-Space Model}{Turney}
\firstpageno{533}
\makeatletter
\newcommand\figcaption{\def\@captype{figure}\caption}
\newcommand\tabcaption{\def\@captype{table}\caption}
\makeatother
\begin{document}

\title{Domain and Function: A Dual-Space Model \\
       of Semantic Relations and Compositions}

\author{\name Peter D. Turney \email peter.turney@nrc-cnrc.gc.ca \\
       \addr National Research Council Canada \\
       Ottawa, Ontario, Canada, K1A 0R6 }

\maketitle

\begin{abstract}
Given appropriate representations of the semantic relations between {\em carpenter} and
{\em wood} and between {\em mason} and {\em stone} (for example, vectors in a vector space
model), a suitable algorithm should be able to recognize that these relations are
highly similar ({\em carpenter} is to {\em wood} as {\em mason} is to {\em stone};
the relations are analogous). Likewise, with representations of {\em dog}, {\em house}, and
{\em kennel}, an algorithm should be able to recognize that the semantic composition
of {\em dog} and {\em house}, {\em dog house}, is highly similar to {\em kennel} ({\em dog house}
and {\em kennel} are synonymous). It seems that these two tasks, recognizing
relations and compositions, are closely connected. However, up to now,
the best models for relations are significantly different from the best models for
compositions. In this paper, we introduce a dual-space model that unifies these
two tasks. This model matches the performance of the best previous models for
relations and compositions. The dual-space model consists of a space
for measuring domain similarity and a space for measuring function similarity.
{\em Carpenter} and {\em wood} share the same domain, the domain of {\em carpentry}.
{\em Mason} and {\em stone} share the same domain, the domain of {\em masonry}.
{\em Carpenter} and {\em mason} share the same function, the function of {\em artisans}.
{\em Wood} and {\em stone} share the same function, the function of {\em materials}.
In the composition {\em dog house}, {\em kennel} has some domain overlap with
both {\em dog} and {\em house} (the domains of {\em pets} and {\em buildings}). The function
of {\em kennel} is similar to the function of {\em house} (the function of {\em shelters}).
By combining domain and function similarities in various ways, we can
model relations, compositions, and other aspects of semantics.
\end{abstract}

\section{Introduction}
\label{sec:intro}

The {\em distributional hypothesis} is that words that occur in similar contexts tend to
have similar meanings \cite{harris54,firth57}. Many vector space models (VSMs) of
semantics use a word--context matrix to represent the distribution of words over contexts,
capturing the intuition behind the distributional hypothesis \cite{turney10}. VSMs
have achieved impressive results at the level of individual words \cite{rapp03}, but it is
not clear how to extend them to the level of phrases, sentences, and beyond. For example,
we know how to represent {\em dog} and {\em house} with vectors, but how should we represent
{\em dog house}$\,$?

One approach to representing {\em dog house} is to treat it as a unit, the same way
we handle individual words. We call this the {\em holistic} or {\em noncompositional}
approach to representing phrases. The holistic approach may be suitable for some phrases,
but it does not scale up. With a vocabulary of $N$ individual words, we can have $N^2$
two-word phrases, $N^3$ three-word phrases, and so on. Even with a very large corpus of
text, most of these possible phrases will never appear in the corpus. People are continually
inventing new phrases, and we are able to understand these new phrases although we have never
heard them before; we are able to infer the meaning of a new phrase by {\em composition} of
the meanings of the component words. This scaling problem could be viewed as an issue of data
sparsity, but it is better to think of it as a problem of {\em linguistic creativity}
\cite{chomsky75,fodor02}. To master natural language, algorithms must be able to represent
phrases by composing representations of individual words. We cannot treat all $n$-grams
($n > 1$) the way we treat unigrams (individual words). On the other hand, the holistic
approach is ideal for idiomatic expressions (e.g., {\em kick the bucket}) for which the
meaning cannot be inferred from the component words.

The creativity and novelty of natural language require us to take a compositional
approach to the majority of the $n$-grams that we encounter. Suppose we have
vector representations of {\em dog} and {\em house}. How can we compose these representations
to represent {\em dog house}$\,$? One strategy is to represent {\em dog house} by the average
of the vectors for {\em dog} and {\em house} \cite{landauer97}. This simple proposal actually
works, to a limited degree \cite{mitchell08,mitchell10}. However {\em boat house}
and {\em house boat} would be represented by the same average vector, yet
they have different meanings. Composition by averaging does not deal with the
{\em order sensitivity} of phrase meaning. \citeA{landauer02} estimates that 80\% of
the meaning of English text comes from word choice and the remaining 20\% comes from
word order.

Similar issues arise with the representation of semantic relations. Given vectors for
{\em carpenter} and {\em wood}, how can we represent the semantic relations between
{\em carpenter} and {\em wood}$\,$? We can treat {\em carpenter}$\,:${\em wood} as a
unit and search for paraphrases of the relations between {\em carpenter} and {\em wood}
\cite{turney06b}. In a large corpus, we could find phrases such as {\em the carpenter cut
the wood}, {\em the carpenter used the wood}, and {\em wood for the carpenter}.
This variation of the holistic approach can enable us to recognize that the semantic
relations between {\em carpenter} and {\em wood} are highly similar to the relations
between {\em mason} and {\em stone}. However, the holistic approach to semantic
relations suffers from the same data sparsity and linguistic creativity problems as
the holistic approach to semantic composition.

We could represent the relation between {\em carpenter} and {\em wood} by averaging
their vectors. This might enable us to recognize that {\em carpenter} is to {\em wood}
as {\em mason} is to {\em stone}, but it would incorrectly suggest that
{\em carpenter} is to {\em wood} as {\em stone} is to {\em mason}. The problem
of order sensitivity arises with semantic relations just as it arose with semantic composition.

Many ideas have been proposed for composing vectors \cite{landauer97,kintsch01,mitchell10}.
\citeA{erk08} point out two problems that are common to several of these proposals.
First, often they do not have the {\em adaptive capacity} to represent the variety
of possible syntactic relations in a phrase. For example, in the phrase {\em a horse draws},
{\em horse} is the subject of the verb {\em draws}, whereas it is the object of the verb
in the phrase {\em draws a horse}. The composition of the vectors for {\em horse} and {\em draws}
must be able to adapt to a variety of syntactic contexts in order to properly model the
given phrases. Second, a single vector is too weak to handle a long phrase, a sentence, or a
document. A single vector ``can only encode a fixed amount of structural information if its
dimensionality is fixed, but there is no upper limit on sentence length, and hence on the
amount of structure to be encoded'' \cite[p.~898]{erk08}. A fixed dimensionality does not
allow {\em information scalability}.

Simple (unweighted) averaging of vectors lacks {\em adaptive capacity}, because it treats
all kinds of composition in the same way; it does not have the flexibility to represent
different modes of composition. A good model must have the capacity to adapt to different
situations. For example, with weighted averaging, the weights
can be tuned for different syntactic contexts \cite{mitchell08,mitchell10}.

{\em Information scalability} means that the size of semantic representations should
grow in proportion to the amount of information that they are representing.
If the size of the representation is fixed, eventually there will be information loss.
On the other hand, the size of representations should not grow exponentially.

One case where the problem of information scalability arises is with approaches that map
multiple vectors into a single vector. For example, if we represent {\em dog house}
by adding the vectors for {\em dog} and {\em house} (mapping two vectors into one),
there may be information loss. As we increase the number of vectors that are mapped
into a single vector, we will eventually reach a point where the single vector
can no longer contain the information from the multiple vectors. This problem
can be avoided if we do not try to map multiple vectors into a single vector.

Suppose we have a $k$-dimensional vector with floating point elements of $b$ bits each.
Such a vector can hold at most $kb$ bits of information. Even if we allow $b$
to grow, if $k$ is fixed, we will eventually have information loss. In a vector space model
of semantics, the vectors have some resistance to noise. If we perturb a vector with noise
below some threshold $\epsilon$, there is no significant change in the meaning that it represents.
Therefore we should think of the vector as a hypersphere with a radius of $\epsilon$, rather than a
point. We may also put bounds $[-r, +r]$ on the range of the values of the elements in the
vector.\footnote{In models where the vectors are normalized to unit length (e.g., models
that use cosine to measure similarity), the elements must lie within the range $[-1, +1]$.
If any element is outside this range, then the length of the vector will be greater than one.
In general, floating point representations have minimum and maximum values.} There is a finite
number $N$ of hyperspheres of radius $\epsilon$ that can be packed into a bounded
$k$-dimensional space \cite{conway98}. According to information theory, if we have a
finite set of $N$ messages, then we need at most $\log_2(N)$ bits to encode a message.
Likewise, if we have a finite set of $N$ vectors, then a vector represents at most
$\log_2(N)$ bits of information. Therefore the information capacity of a single
vector in bounded $k$-dimensional space is limited to $\log_2(N)$ bits.

Past work suggests that recognizing relations and compositions are closely
connected tasks \cite{kintsch00,kintsch01,mangalath04}. The goal of our research
is a unified model that can handle both compositions and relations,
while also resolving the issues of linguistic creativity, order sensitivity,
adaptive capacity, and information scalability. These considerations have led us to
a dual-space model, consisting of a domain space for measuring
domain similarity (i.e., topic, subject, or field similarity) and a function space for
measuring function similarity (i.e., role, relationship, or usage similarity).

In an analogy $a\!:\!b\!::\!c\!:\!d$ ($a$ is to $b$ as $c$ is to $d$;
for example, {\em traffic} is to {\em street} as {\em water} is to {\em riverbed}),
$a$ and $b$ have relatively high domain similarity ({\em traffic} and {\em street} come
from the domain of {\em transportation}) and $c$ and $d$ have relatively high domain
similarity ({\em water} and {\em riverbed} come from the domain of {\em hydrology}).
On the other hand, $a$ and $c$ have relatively high function similarity ({\em traffic}
and {\em water} have similar roles in their respective domains; they are both things that
{\em flow}) and $b$ and $d$ have relatively high function similarity ({\em street}
and {\em riverbed} have similar roles in their respective domains; they are both
things that {\em carry} things that flow). By combining domain and function
similarity in appropriate ways, we can recognize that the semantic relations
between  {\em traffic} and {\em street} are analogous to the relations
between {\em water} and {\em riverbed}.

For semantic composition, the appropriate way to combine similarities may
depend on the syntax of the composition. Let's focus on noun-modifier composition
as an example. In the noun-modifier phrase $ab$ (for instance, {\em brain doctor}),
the head noun $b$ ({\em doctor}) is modified by an adjective or noun $a$ ({\em brain}).
Suppose we have a word $c$ ({\em neurologist}) that is synonymous with $ab$.
The functional role of the noun-modifier phrase $ab$ is determined by the
head noun $b$ (a {\em brain doctor} is a kind of {\em doctor}) and $b$
has a relatively high degree of function similarity with $c$ ({\em doctor} and
{\em neurologist} both function as {\em doctors}). Both $a$ and $b$ have a high
degree of domain similarity with $c$ ({\em brain}, {\em doctor}, and {\em neurologist}
all come from the domain of {\em clinical neurology}). By combining domain and function
similarity, we can recognize that {\em brain doctor} is synonymous
with {\em neurologist}.

Briefly, the proposal is to compose similarity measures instead of composing vectors.
That is, we apply various mathematical functions to combine cosine similarity
measures, instead of applying the functions directly to the vectors. This addresses the
information loss problem, because we preserve the vectors for the individual component words.
(We do not map multiple vectors into a single vector.) Since we have two different
spaces, we also have flexibility to address the problem of adaptive capacity.\footnote{Two
similarity spaces give us more options for similarity composition than one space, just as
two types of characters (0 and 1) give us more options for generating
strings than one type of character (0 alone).} This model is compositional, so it resolves
the linguistic creativity problem. We deal with order sensitivity by combining similarity
measures in ways that recognize the effects of word order.

It might be argued that what we present here is {\em not} a model of semantic composition,
but a way to compare the words that form two phrases in order to derive a measure
of similarity of the phrases. For example, in Section~\ref{subsec:phrase-exper}
we derive a measure of similarity for the phrases {\em environment secretary} and
{\em defence minister}, but we do not actually provide a representation for the
phrase {\em environment secretary}. On the other hand, most past work on the problem
of semantic composition (reviewed in Section~\ref{subsec:composition}) yields
a representation for the composite phrase {\em environment secretary} that is
different from the union of the representations of the component words, {\em environment}
and {\em secretary}.

This argument is based on the assumption that the goal of semantic composition
is to create a single, general-purpose, stand-alone representation of a phrase,
as a composite, distinct from the union of the representations of the component words.
This assumption is not necessary and our approach does not use this assumption.
We believe that this assumption has held back progress on the problem of semantic
composition.

We argue that what we present here {\em is} a model of semantic composition, but
it is composition of similarities, not composition of vectors. Vectors
can represent individual words, but similarities inherently represent
relations between two (or more) things. Composing vectors can yield
a stand-alone representation of a phrase, but composing similarities
necessarily yields a linking structure that connects a phrase to other phrases.
Similarity composition does not result in a stand-alone representation of a phrase,
but practical applications do not require stand-alone representations.
Whatever practical tasks can be performed with stand-alone representations of phrases,
we believe can be performed equally well (or better) with similarity composition.
We discuss this issue in more depth in Section~\ref{sec:theory}.

The next section surveys related work on the modeling of semantic composition
and semantic relations. Section~\ref{sec:spaces} describes how we build {\em domain} and
{\em function} space. To test the hypothesis that there is value in having two separate
spaces, we also create {\em mono} space, which is the merger of the domain and function spaces.
We then present four sets of experiments with the dual-space model in
Section~\ref{sec:experiments}. We evaluate the dual-space approach with multiple-choice
analogy questions from the SAT \cite{turney06b}, multiple-choice noun-modifier composition
questions derived from WordNet \cite{fellbaum98}, phrase similarity rating problems
\cite{mitchell10}, and similarity versus association problems 
\cite{chiarello90}. We discuss the experimental results in Section~\ref{sec:discussion}.
Section~\ref{sec:theory} considers some theoretical questions about the dual-space model.
Limitations of the model are examined in Section~\ref{sec:future}.
Section~\ref{sec:conclusions} concludes.

This paper assumes some familiarity with vector space models of semantics. For an overview
of semantic VSMs, see the papers in the {\em Handbook of Latent Semantic Analysis}
\cite{landauer07}, the review in \citeS{mitchell10} paper, or the survey by
\citeA{turney10}.

\section{Related Work}
\label{sec:related}

Here we examine related work with semantic composition and relations. In the
introduction, we mentioned four problems with semantic models, which yield
four desiderata for a semantic model:


\begin{enumerate}[itemsep=2pt,parsep=1pt,topsep=4pt,partopsep=1pt]

\item {\bf Linguistic creativity:} The model should be able to handle phrases (in the case
of semantic composition) or word pairs (in the case of semantic relations) that it
has never seen before, when it is familiar with the component words.

\item {\bf Order sensitivity:} The model should be sensitive to the order of the words
in a phrase (for composition) or a word pair (for relations), when the order
affects the meaning.

\item {\bf Adaptive capacity:} For phrases, the model should have the flexibility to
represent different kinds of syntactic relations. For word pairs, the model should have the
flexibility to handle a variety of tasks, such as measuring the degree of
{\em relational} similarity between two pairs (see Section~\ref{subsec:sat-exper})
versus measuring the degree of {\em phrasal} similarity between two pairs
(see Section~\ref{subsec:phrase-exper}).

\item {\bf Information scalability:} For phrases, the model should scale up
with neither loss of information nor exponential growth in representation size
as the number of component words in the phrases increases. For $n$-ary semantic
relations \cite{turney08b}, the model should scale up with neither loss of information
nor exponential growth in representation size as $n$, the number of terms in the relations,
increases.

\end{enumerate}

\noindent We will review past work in the light of these four considerations.

\subsection{Semantic Composition}
\label{subsec:composition}

Let $ab$ be a phrase, such as a noun-modifier phrase, and assume that we have vectors
$\mathbf{a}$ and $\mathbf{b}$ that represent the component words $a$ and $b$. One of the
earliest proposals for semantic composition is to represent $ab$ by the vector $\mathbf{c}$
that is the average of $\mathbf{a}$ and $\mathbf{b}$ \cite{landauer97}. If we are using a
cosine measure of vector similarity, taking the average of a set of vectors (or
their centroid) is the same as adding the vectors, $\mathbf{c} = \mathbf{a} + \mathbf{b}$.
Vector addition works relatively well in practice \cite{mitchell08,mitchell10}, although it
lacks order sensitivity, adaptive capacity, and information scalability. Regarding order
sensitivity and adaptive capacity, \citeA{mitchell08,mitchell10} suggest using weights,
$\mathbf{c} = \alpha \mathbf{a} + \beta \mathbf{b}$, and tuning the weights
to different values for different syntactic relations. In their experiments
\cite{mitchell10}, weighted addition performed better than unweighted addition.

\citeA{kintsch01} proposes a variation of additive composition in which $\mathbf{c}$
is the sum of $\mathbf{a}$, $\mathbf{b}$, and selected neighbours $\mathbf{n}_i$ of
$\mathbf{a}$ and $\mathbf{b}$, $\mathbf{c} = \mathbf{a} + \mathbf{b} + \sum_i \mathbf{n}_i$.
The neighbours are vectors for other words in the given vocabulary
(i.e., other rows in the given word--context matrix). The neighbours are chosen in
a manner that attempts to address order sensitivity and adaptive capacity, but there
is still a problem with information scalability due to fixed dimensionality.
\citeA{utsumi09} presents a similar model, but with a different way of selecting neighbours.
\citeA{mitchell10} found that a simple additive model peformed better than an additive model
that included neighbours.

\citeA{mitchell08,mitchell10} suggest element-wise multiplication as a
composition operation, $\mathbf{c} = \mathbf{a} \odot \mathbf{b}$, where
$c_i = a_i \, \cdot \, b_i$. Like vector addition, element-wise multiplication
suffers from a lack of order sensitivity, adaptive capacity, and information scalability.
Nonetheless, in an experimental evaluation of seven compositional models and two
noncompositional models, element-wise multiplication had the best performance
\cite{mitchell10}.

Another approach is to use a tensor product for composition
\cite{smolenksy90,aerts04,clark07,widdows08},
such as the outer product, $\mathbf{C} = \mathbf{a} \otimes \mathbf{b}$.
The outer product of two vectors ($\mathbf{a}$ and $\mathbf{b}$), each  with $n$
elements, is an $n \times n$ matrix ($\mathbf{C}$). The outer product of three
vectors is an $n \times n \times n$ third-order tensor. This results in an information
scalability problem: The representations grow exponentially large as the phrases
grow longer.\footnote{There are ways to avoid the exponential growth;
for example, a third-order tensor with a rank of 1 on all three modes may be compactly
encoded by its three component vectors. \citeA{kolda09} discuss compact tensor representations.}
Furthermore, the outer product did not perform as well as element-wise multiplication in
\citeS{mitchell10} experiments. Recent work with tensor products \cite{clark08,grefenstette11}
has attempted to address the issue of information scalability.

Circular convolution is similar to the outer product, but the outer product matrix is
compressed back down to a vector, $\mathbf{c} = \mathbf{a} \, \circledast \, \mathbf{b}$
\cite{plate95,jones07}. This avoids information explosion, but it results in information loss.
Circular convolution performed poorly in \citeS{mitchell10} experiments.

\citeA{baroni10} and \citeA{guevara10} suggest another model of composition for
adjective-noun phrases. The core strategy that they share is to use a few holistic vectors
to train a compositional model. With partial least squares regression (PLSR), we can
learn a linear model that maps the vectors for the component nouns and adjectives
to linear approximations of the holistic vectors for the phrases. The linguistic creativity problem
is avoided because the linear model only needs a few holistic vectors for training; there is
no need to have holistic vectors for all plausible adjective-noun phrases. Given a phrase
that is not in the training data, the linear model predicts the holistic vector
for the phrase, given the component vectors for the adjective and the noun. This 
works well for adjective-noun phrases, but it is not clear how to generalize it to
other parts of speech or to longer phrases.

One application for semantic composition is measuring the similarity of phrases
\cite{erk08,mitchell10}. Kernel methods have been applied to the closely related
task of identifying paraphrases \cite{moschitti08}, but the emphasis with
kernel methods is on syntactic similarity, rather than semantic similarity.

Neural network models have been combined with vector space models for the task of
language modeling \cite{bengio03,socher10,socher11}, with impressive results.
The goal of a language model is to estimate the probability of a phrase or
to decide which of several phrases is the most likely. VSMs can improve the
probability estimates of a language model by measuring the similarity
of the words in the phrases and smoothing probabilities over groups of
similar words. However, in a language model, words are considered similar to the
degree that they can be exchanged without altering the {\em probability} of a
given phrase, without regard to whether the exchange alters the {\em meaning} of
the phrase. This is like function similarity, which measures the
degree to which words have similar functional roles, but these language
models are missing anything like domain similarity.

\citeA{erk08} present a model that is similar to ours in that it
has two parts, a vector space for measuring similarity and a model of selectional
preferences. Their vector space is similar to domain space and their
model of selectional preferences plays a role similar to function space.
An individual word $a$ is represented by a triple, $A = \langle \mathbf{a}, R, R^{-1}\rangle$,
consisting of the word's vector, $\mathbf{a}$, its selectional preferences,
$R$, and its inverse selectional preferences, $R^{-1}$. A phrase $ab$ is
represented by a pair of triples, $\langle A', B'\rangle$.
The triple $A'$ is a modified form of the triple $A$ that represents the
individual word $a$. The modifications adjust the representation to model
how the meaning of $a$ is altered by its relation to $b$ in the phrase $ab$.
Likewise, the triple $B'$ is a modified form of the triple $B$ that represents $b$,
such that $B'$ takes into account how $a$ affects $b$.

When $A$ is transformed to $A'$ to represent the influence of $b$ on the
meaning of $a$, the vector $\mathbf{a}$ in $A$ is transformed to a new
vector $\mathbf{a}'$ in $A'$. Let $\mathbf{r}_b$ be a vector that represents
the typical words that are consistent with the selectional preferences of $b$.
The vector $\mathbf{a}'$ is the composition of $\mathbf{a}$ with $\mathbf{r}_b$.
\citeA{erk08} use element-wise multiplication for composition,
$\mathbf{a}' = \mathbf{a} \odot \mathbf{r}_b$. The intention is to make $\mathbf{a}$
more like a typical vector $\mathbf{x}$ that would be expected for a phrase $xb$.
Likewise, for $\mathbf{b}'$ in $B'$, we have $\mathbf{b}' = \mathbf{b} \odot \mathbf{r}_a$

\citeS{erk08} model and related models \cite{thater10} address linguistic
creativity, order sensitivity, adaptive capacity, and information scalability,
but they are not suitable for measuring the similarity of semantic relations.
Consider the analogy {\em traffic} is to {\em street} as {\em water} is to {\em riverbed}.
Let $\langle A', B'\rangle$ represent {\em traffic}$\,:${\em street} and let
$\langle C', D'\rangle$ represent {\em water}$\,:${\em riverbed}. The transformation
of $A$, $B$, $C$, and $D$ to $A'$, $B'$, $C'$, and $D'$ reinforces the connection
between {\em traffic} and {\em street} and between {\em water} and {\em riverbed},
but it does not help us recognize the relational similarity between
{\em traffic}$\,:${\em street} and {\em water}$\,:${\em riverbed}.
Of course, these models were not designed for relational similarity, so this
is not surprising. However, the goal here is to find a unified model that
can handle both compositions and relations.

\subsection{Semantic Relations}
\label{subsec:relations}

For semantic relations, we can make some general observations about order sensitivity.
Let $a\!:\!b$ and $c\!:\!d$ be two word pairs and let
${\rm sim_r}(a\!:\!b, c\!:\!d) \in \Re$ be a measure of the degree of similarity
between the relations of $a\!:\!b$ and $c\!:\!d$. If $a\!:\!b\!::\!c\!:\!d$
is a good analogy, then ${\rm sim_r}(a\!:\!b, c\!:\!d)$ will have a relatively
high value. In general, a good model of relational similarity should respect the
following equalities and inequalities:

{\allowdisplaybreaks 
\begin{align}
\label{eqn:simr1} {\rm sim_r}(a\!:\!b, c\!:\!d) & = {\rm sim_r}(b\!:\!a, d\!:\!c) \\
\label{eqn:simr2} {\rm sim_r}(a\!:\!b, c\!:\!d) & = {\rm sim_r}(c\!:\!d, a\!:\!b) \\
\label{eqn:simr3} {\rm sim_r}(a\!:\!b, c\!:\!d) & \neq {\rm sim_r}(a\!:\!b, d\!:\!c) \\
\label{eqn:simr4} {\rm sim_r}(a\!:\!b, c\!:\!d) & \neq {\rm sim_r}(a\!:\!d, c\!:\!b)
\end{align}
} 

\noindent For example, given that {\em carpenter}$\,:${\em wood} and
{\em mason}$\,:${\em stone} make a good analogy, it follows from Equation~\ref{eqn:simr1}
that {\em wood}$\,:${\em carpenter} and {\em stone}$\,:${\em mason} make an
equally good analogy. Also, according to Equation~\ref{eqn:simr2}, {\em mason}$\,:${\em stone}
and {\em carpenter}$\,:${\em wood} make a good analogy. On the other hand, as suggested by
Equation~\ref{eqn:simr3}, {\em carpenter}$\,:${\em wood} is {\em not} analogous to
{\em stone}$\,:${\em mason}. Likewise, as indicated by Equation~\ref{eqn:simr4}, it is a
{\em poor} analogy to assert that {\em carpenter} is to {\em stone} as {\em mason} is
to {\em wood}.

\citeA{rosario01} present an algorithm for classifying word pairs according
to their semantic relations. They use a lexical hierarchy to map word
pairs to feature vectors. Any classification scheme implicitly tell us
something about similarity. Two word pairs that are in the same semantic relation
class are implicitly more relationally similar than two word pairs in different classes.
When we consider the relational similarity that is implied by \citeS{rosario01} algorithm,
we see that there is a problem of order sensitivity: Equation~\ref{eqn:simr4} is violated.

Let ${\rm sim_h}(x,y) \in \Re$ be a measure of the degree of hierarchical similarity
between the words $x$ and $y$. If ${\rm sim_h}(x,y)$ is relatively high,
then $x$ and $y$ share a common hypernym relatively close to them in the given
lexical hierarchy. In essence, the intuition behind \citeS{rosario01} algorithm is, if
both ${\rm sim_h}(a,c)$ and ${\rm sim_h}(b,d)$ are high, then
${\rm sim_r}(a\!:\!b, c\!:\!d)$ should also be high. That is, if
${\rm sim_h}(a,c)$ and ${\rm sim_h}(b,d)$ are high enough, then
$a\!:\!b$ and $c\!:\!d$ should be assigned to the same relation class.

For example, consider the analogy {\em mason} is to {\em stone} as {\em carpenter}
is to {\em wood}. The common hypernym of {\em mason} and {\em carpenter} is
{\em artisan}; we can see that ${\rm sim_h}(mason,carpenter)$ is high. The common
hypernym of {\em stone} and {\em wood} is {\em material}; hence ${\rm sim_h}(stone,wood)$
is high. It seems that a good analogy is indeed characterized by high values for
${\rm sim_h}(a,c)$ and ${\rm sim_h}(b,d)$. However, the symmetry of ${\rm sim_h}(x,y)$
leads to a problem. If ${\rm sim_h}(b,d)$ is high, then ${\rm sim_h}(d,b)$ must
also be high, but this implies that ${\rm sim_r}(a\!:\!d, c\!:\!b)$ is high.
That is, we incorrectly conclude that {\em mason} is to {\em wood}
as {\em carpenter} is to {\em stone} (see Equation~\ref{eqn:simr4}).

Some later work with classifying semantic relations has used different
algorithms, but the same underlying intuition about hierarchical similarity
\cite{rosario02,nastase03,nastase06}. We use a similar intuition here, since
similarity in function space is closely related to hierarchical similarity,
${\rm sim_h}(x,y)$, as we will see later (Section~\ref{subsec:chiarello-exper}).
However, including domain space in the relational similarity measure saves us
from violating Equation~\ref{eqn:simr4}.

Let ${\rm sim_f}(x,y) \in \Re$ be function similarity as measured by
the cosine of vectors $\mathbf{x}$ and $\mathbf{y}$ in function space. Let
${\rm sim_d}(x,y) \in \Re$ be domain similarity as measured by
the cosine of vectors $\mathbf{x}$ and $\mathbf{y}$ in domain space.
Like past researchers \cite{rosario01,rosario02,nastase03,veale04,nastase06},
we look for high values of ${\rm sim_f}(a,c)$ and ${\rm sim_f}(b,d)$
as indicators that ${\rm sim_r}(a\!:\!b, c\!:\!d)$ should be high, but we
also look for high values of ${\rm sim_d}(a,b)$ and ${\rm sim_d}(c,d)$.
Continuing the previous example, we do {\em not} conclude
that {\em mason} is to {\em wood} as {\em carpenter} is to {\em stone},
because {\em wood} does not belong in the domain of {\em masonry} and
{\em stone} does not belong in the domain of {\em carpentry}.

Let D be a determiner (e.g., {\em the}, {\em a}, {\em an}).
\citeA{hearst92} showed how patterns of the form ``D X such as D Y''
(``a bird such as a crow'') or ``D Y is a kind of X'' (``the crow is a kind
of bird'') can be used to infer that X is a hypernym of Y ({\em bird}
is a hypernym of {\em crow}). A pair--pattern matrix is a VSM
in which the rows are word pairs and the columns are various
``X \dots Y'' patterns. \citeA{turney03b} demonstrated that a pair--pattern VSM
can be used to measure relational similarity. Suppose we have a pair-pattern
matrix $\mathbf{X}$ in which the word pair $a\!:\!b$ corresponds to the row
vector $\mathbf{x}_{i}$ and $c\!:\!d$ corresponds to $\mathbf{x}_{j}$. The
approach is to measure the relational similarity ${\rm sim_r}(a\!:\!b, c\!:\!d)$
by the cosine of $\mathbf{x}_{i}$ and $\mathbf{x}_{j}$.

At first the patterns in these pair--pattern matrices were generated by hand
\cite{turney03b,turney05a}, but later work \cite{turney06b} used automatically
generated patterns. Other authors have used variations of this technique
\cite{nakov06,nakov07,davidov08,bollegala09,oseaghdha09}. All of these models suffer
from the linguistic creativity problem. Because the models are noncompositional (holistic),
they cannot scale up to handle the huge number of possible pairs.
Even the largest corpus cannot contain all the pairs that a human speaker
might use in daily conversation.

\citeA{turney06b} attempted to handle the linguistic creativity problem within a holistic model
by using synonyms. For example, if a corpus does not contain {\em traffic} and {\em street}
within a certain window of text, perhaps it might contain {\em traffic} and {\em road}.
If it does not contain {\em water} and {\em riverbed}, perhaps it has {\em water} and
{\em channel}. However, this is at best a partial solution. \citeS{turney06b} algorithm
required nine days to process 374 multiple choice SAT analogy questions. Using the
dual-space model, without specifying in advance what word pairs it might face,
we can answer the 374 questions in a few seconds (see Section~\ref{subsec:sat-exper}).
Compositional models scale up better than holistic models.

\citeA{mangalath04} presented a model for semantic relations that represents
word pairs with vectors of ten abstract relational categories, such as {\em hyponymy},
{\em meronymy}, {\em taxonomy}, and {\em degree}. The approach is to construct a kind of
second-order vector space in which the elements of the vectors are degrees
of similarity, calculated from cosines with a first-order word--context matrix.

For instance, {\em carpenter}$\,:${\em wood} can be represented by a second-order
vector composed of ten cosines calculated from first-order vectors. In this second-order
vector, the value of the element corresponding to, say, {\em meronymy} would be
the cosine of two first-order vectors, $\mathbf{x}$ and $\mathbf{y}$. The
vector $\mathbf{x}$ would be the sum of the first-order vectors for {\em carpenter}
and {\em wood}. The vector $\mathbf{y}$ would be the sum of several vectors
for words that are related to {\em meronymy}, such as {\em part}, {\em whole},
{\em component}, {\em portion}, {\em contains}, {\em constituent}, and {\em segment}.
The cosine of $\mathbf{x}$ and $\mathbf{y}$ would indicate the degree to which
{\em carpenter} and {\em wood} are related to {\em meronymy}.

\citeS{mangalath04} model suffers from information scalability and order sensitivity problems.
Information loss takes place when the first-order vectors are summed and also when the
high-dimensional first-order space is reduced to a ten-dimensional second-order space.
The order sensitivity problem is that the second-order vectors violate Equation~\ref{eqn:simr3},
because the pairs $c\!:\!d$ and $d\!:\!c$ are represented by the same second-order
vector.

A natural proposal is to represent a word pair $a\!:\!b$ the same way we would
represent a phrase $ab$. That is, whatever compositional model we have for
phrases could also be applied to word pairs. However any problems that the
compositional model has with order sensitivity or information scalability carry over to word pairs.
For example, if we represent $a\!:\!b$ by $\mathbf{c} = \mathbf{a} + \mathbf{b}$
or $\mathbf{c} = \mathbf{a} \odot \mathbf{b}$, then we violate Equation~\ref{eqn:simr3},
because $\mathbf{a} + \mathbf{b} =\mathbf{b} + \mathbf{a}$ and
$\mathbf{a} \odot \mathbf{b} =\mathbf{b} \odot \mathbf{a}$.

\section{Three Vector Spaces}
\label{sec:spaces}

In this section, we describe three vector space models. All three spaces consist of
word--context matrices, in which the rows correspond to words and the columns correspond
to the contexts in which the words occur. The differences among the three spaces
are in the kinds of contexts. Domain space uses nouns for context, function space
uses verb-based patterns for context, and mono space is a merger of the domain
and function contexts. Mono space was created in order to test the hypothesis that it is
useful to separate the domain and function spaces; mono space serves as a baseline.

\subsection{Constructing the Word--Context Matrices}
\label{subsec:building}

Building the three spaces involves a series of steps. There are three main steps,
each of which has a few substeps. The first and last steps are the
same for all three spaces; the differences in the spaces are the result of differences
in the second step.


\begin{enumerate}[itemsep=4pt,parsep=1pt,topsep=4pt,partopsep=1pt]

\item {\bf Find terms in contexts:} input: a corpus and a lexicon, output: terms in contexts.

\begin{enumerate}[itemsep=1pt,parsep=1pt,topsep=4pt,partopsep=1pt,label*=\arabic*.]

\item Extract terms from the lexicon and find their frequencies in the corpus.

\item Select all terms above a given frequency as candidate rows for the frequency matrix.

\item For each selected term, find phrases in the corpus that contain the term within
a given window size.

\item Use a tokenizer to split the phrases into tokens.

\item Use a part-of-speech tagger to tag the tokens in the phrases.

\end{enumerate}

\item  {\bf Build a term--context frequency matrix:} input: terms in contexts, output:
a sparse frequency matrix.

\begin{enumerate}[itemsep=1pt,parsep=1pt,topsep=4pt,partopsep=1pt,label*=\arabic*.]

\item Convert the tagged phrases into contextual patterns (candidate columns).

\item For each contextual pattern, count the number of terms (candidate rows) that
generated the pattern and rank the patterns in descending order of their counts.

\item Select the top $n_c$ contextual patterns as the columns of the matrix.

\item From the initial set of rows (from Step 1.2), drop any row that does not
match any of the top $n_c$ contextual patterns, yielding the final set of $n_r$ rows.

\item For each row (term) and each column (contextual pattern), count the number
of phrases (from Step 1.5) containing the given term and matching the given pattern,
and output the resulting numbers as a sparse frequency matrix.

\end{enumerate}

\item  {\bf Weight the elements and smooth the matrix:} input: a sparse frequency matrix,
output: the singular value decomposition (SVD) of the weighted matrix.

\begin{enumerate}[itemsep=1pt,parsep=1pt,topsep=4pt,partopsep=1pt,label*=\arabic*.]

\item Convert the raw frequencies to positive pointwise mutual information (PPMI) values.

\item Apply SVD to the PPMI matrix and output the SVD component matrices.

\end{enumerate}
\end{enumerate}

The input corpus in Step 1 is a collection of web pages gathered from university websites
by a webcrawler.\footnote{The corpus was collected by Charles Clarke at the University of
Waterloo.} The corpus contains approximately $5 \times 10^{10}$ words, which comes to about 280
gigabytes of plain text. To facilitate finding term frequencies and sample phrases,
we indexed this corpus with the Wumpus search engine \cite{buettcher05}.\footnote{Wumpus
is available at http://www.wumpus-search.org/.} The rows for the matrices were
selected from terms (words and phrases) in the WordNet lexicon.\footnote{WordNet is
available at http://wordnet.princeton.edu/.} We found that selecting terms from WordNet
resulted in subjectively higher quality than simply selecting terms with high corpus frequencies.

In Step 1.1, we extract all unique words and phrases ($n$-grams) from the {\em index.sense}
file in WordNet 3.0, skipping $n$-grams that contain numbers (only letters, hyphens, and
spaces are allowed in the $n$-grams). We find the $n$-gram corpus frequencies by querying
Wumpus with each $n$-gram. All $n$-grams with a frequency of at least 100 and at least
2 characters are candidate rows in Step 1.2. For each selected $n$-gram, we query Wumpus to
find a maximum of 10,000 phrases in Step 1.3.\footnote{The limit of 10,000 phrases per $n$-gram
is required to make Wumpus run in a tolerable amount of time. Finding phrases is the most
time-consuming step in the construction of the spaces. We use a solid-state drive (SSD) to speed
up this step.} The phrases are limited to a window of 7 words to the left of the $n$-gram and 7
words to the right, for a total window size of $14+n$ words. We use OpenNLP 1.3.0 to tokenize and
part-of-speech tag the phrases (Steps 1.4 and 1.5).\footnote{OpenNLP is available at
http://incubator.apache.org/opennlp/.} The tagged phrases come to about 46 gigabytes.\footnote{The
tagged phrases are available from the author on request.}

In Step 2.1, we generate contextual patterns from the part-of-speech tagged phrases.
Different kinds of patterns are created for the three different kinds of spaces. The
details of this step are given in the following subsections. Each phrase may yield
several patterns. The three spaces each have more than 100,000 rows, with a maximum of
10,000 phrases per row and several patterns per phrase. This can result in millions of
distinct patterns, so we filter the patterns in Steps 2.2 and 2.3. We select the
top $n_c$ patterns that are shared by the largest number of rows. Given the large number
of patterns, they may not all fit in RAM. To work with limited RAM, we use the Linux
{\em sort} command, which is designed to efficiently sort files that are too large
to fit in RAM. For each row, we make a file of the distinct patterns generated by
that row. We then concatenate all of the files for all of the rows and alphabetically sort
the patterns in the concatenated file. In the sorted file, identical patterns are
adjacent, which makes it easy to count the number of occurrences of each pattern.
After counting, a second sort operation yields a ranked list of patterns, from which
we select the top $n_c$.

It is possible that some of the candidate rows from Step 1.2 might not match any of
the patterns from Step 2.3. These rows would be all zeros in the matrix, so we
remove them in Step 2.4. Finally, we output a sparse frequency matrix $\mathbf{F}$
with $n_r$ rows and $n_c$ columns. If the $i$-th row corresponds to the $n$-gram $w_i$
and the $j$-th column corresponds to the contextual pattern $c_j$, then the value of
the element $f_{ij}$ in $\mathbf{F}$ is the number of phrases containing $w_i$ (from
Step 1.5) that generate the pattern $c_j$ (in Step 2.1). In Step 3.2, we use SVDLIBC 1.34
to calculate the singular value decomposition, so the format of the output sparse
matrix in Step 2.5 is chosen to meet the requirements of SVDLIBC.\footnote{SVDLIBC is available
at http://tedlab.mit.edu/$\scriptstyle\sim$dr/svdlibc/.}

In Step 3.1, we apply positive pointwise mutual information (PPMI) to the sparse
frequency matrix $\mathbf{F}$. This is a variation of pointwise mutual information
(PMI) \cite{church89,turney01} in which all PMI values that are less than zero are
replaced with zero \cite{niwa94,bullinaria07}. Let $\mathbf{X}$ be the matrix that
results when PPMI is applied to $\mathbf{F}$. The new matrix $\mathbf{X}$ has the same
number of rows and columns as the raw frequency matrix $\mathbf{F}$. The value of an
element $x_{ij}$ in $\mathbf{X}$ is defined as follows:

{\allowdisplaybreaks 
\begin{align}
p_{ij} & = \frac{f_{ij}}{\sum_{i=1}^{n_r} \sum_{j=1}^{n_c} f_{ij}} \\
p_{i*} & = \frac{\sum_{j=1}^{n_c} f_{ij}}{\sum_{i=1}^{n_r} \sum_{j=1}^{n_c} f_{ij}} \\
p_{*j} & = \frac{\sum_{i=1}^{n_r} f_{ij}}{\sum_{i=1}^{n_r} \sum_{j=1}^{n_c} f_{ij}} \\
{\rm pmi}_{ij} & = \log \left ( \frac{p_{ij}}{p_{i*} p_{*j}} \right ) \\
x_{ij} & =
\left\{
\begin{array}{rl}
{\rm pmi}_{ij} & \mbox{if ${\rm pmi}_{ij} > 0$} \\
0 & \mbox{otherwise}
\end{array}
\right.
\end{align}
} 

In this definition, $p_{ij}$ is the estimated probability that the word $w_i$ occurs
in the context $c_j$, $p_{i*}$ is the estimated probability of the word $w_i$, and
$p_{*j}$ is the estimated probability of the context $c_j$. If $w_i$ and $c_j$ are
statistically independent, then $p_{ij} = p_{i*} p_{*j}$ (by the definition of
independence), and thus ${\rm pmi}_{ij}$ is zero (since $\log(1) = 0$). The product
$p_{i*} p_{*j}$ is what we would expect for $p_{ij}$ if $w_i$ occurs in $c_j$ by pure
random chance. On the other hand, if there is an interesting semantic relation between
$w_i$ and $c_j$, then we should expect $p_{ij}$ to be larger than it would be if $w_i$
and $c_j$ were indepedent; hence we should find that $p_{ij} > p_{i*} p_{*j}$, and
thus ${\rm pmi}_{ij}$ is positive. If the word $w_i$ is unrelated to (or incompatible with)
the context $c_j$, we may find that ${\rm pmi}_{ij}$ is negative. PPMI is designed to give a
high value to $x_{ij}$ when there is an interesting semantic relation between $w_i$ and $c_j$;
otherwise, $x_{ij}$ should have a value of zero, indicating that the occurrence of
$w_i$ in $c_j$ is uninformative.

Finally, in Step 3.2, we apply SVDLIBC to $\mathbf{X}$.
SVD decomposes $\mathbf{X}$ into the product of three matrices
$\mathbf{U} \mathbf{\Sigma} \mathbf{V}^\mathsf{T}$,
where $\mathbf{U}$ and $\mathbf{V}$ are in column orthonormal
form (i.e., the columns are orthogonal and have unit length,
$\mathbf{U}^\mathsf{T} \mathbf{U} = \mathbf{V}^\mathsf{T} \mathbf{V} = \mathbf{I}$)
and $\mathbf{\Sigma}$ is a diagonal matrix of singular values \cite{golub96}.
If $\mathbf{X}$ is of rank $r$, then $\mathbf{\Sigma}$ is also of rank $r$.
Let ${\mathbf{\Sigma}}_k$, where $k < r$, be the diagonal matrix formed from the top $k$
singular values, and let $\mathbf{U}_k$ and $\mathbf{V}_k$ be the matrices produced
by selecting the corresponding columns from $\mathbf{U}$ and $\mathbf{V}$. The matrix
$\mathbf{U}_k \mathbf{\Sigma}_k \mathbf{V}_k^\mathsf{T}$ is the matrix of rank $k$
that best approximates the original matrix $\mathbf{X}$, in the sense that it
minimizes the approximation errors. That is,
${\bf \hat X} = \mathbf{U}_k \mathbf{\Sigma}_k \mathbf{V}_k^\mathsf{T}$
minimizes $\| {{\bf \hat X} - \mathbf{X}} \|_F$
over all matrices ${\bf \hat X}$ of rank $k$, where $\| \ldots \|_F$
denotes the Frobenius norm \cite{golub96}. The final output is the three
matrices, $\mathbf{U}_k$, $\mathbf{\Sigma}_k$, and $\mathbf{V}_k$, that form the
truncated SVD, ${\bf \hat X} = \mathbf{U}_k \mathbf{\Sigma}_k \mathbf{V}_k^\mathsf{T}$.

\subsection{Domain Space}
\label{subsec:domain}

The intuition behind domain space is that the domain or topic of a word
is characterized by the nouns that occur near it. We use a relatively wide window
and we ignore the syntactic context in which the nouns appear.

For domain space, in Step 2.1, each tagged phrase generates at most two contextual
patterns. The contextual patterns are simply the first noun to the left of the
given $n$-gram (if there is one) and the first noun to the right (if there is one).
Since the window size is 7 words on each side of the $n$-gram, there are usually
nouns on both sides of the $n$-gram. The nouns may be either common nouns or
proper nouns. OpenNLP uses the Penn Treebank tags \cite{santorini90}, which include
several different categories of noun tags. All of the noun tags begin with a capital N,
so we simply extract the first words to the left and right of the $n$-gram
that have tags that begin with N. The extracted nouns are converted to lower case.
If the same noun appears on both sides of the $n$-gram, only one contextual
pattern is generated. The extracted patterns are always unigrams; in a noun compound,
only the component noun closest to the $n$-gram is extracted.

Table~\ref{tab:dompats} shows some examples for the $n$-gram {\em boat}. Note that
the window of 7 words does not count punctuation, so the number of tokens in the
window may be greater than the number of words in the window. We can see from
Table~\ref{tab:dompats} that the row vector for the $n$-gram {\em boat}
in the frequency matrix $\mathbf{F}$ will have nonzero values (for example) in the
columns for {\em lake} and {\em summer} (assuming that these contextual patterns
make it through the filtering in Step 2.3).

\begin{table}[h]
\begin{center}
\scalebox{0.92}{
\begin{tabular}{p{0.1in} p{5in} p{0.7in}}
\hline
& Tagged phrases & Patterns \\
\hline
1 & ``would/MD visit/VB Big/NNP Lake/NNP and/CC take/VB our/PRP\$ {\bf boat/NN} on/IN
this/DT huge/JJ beautiful/JJ lake/NN ./. There/EX was/VBD'' \newline & ``lake'' \\[-6pt]
2 & ``the/DT large/JJ paved/JJ parking/NN lot/NN in/IN the/DT {\bf boat/NN} ramp/NN
area/NN and/CC walk/VB south/RB along/IN the/DT'' \newline & ``lot'' \newline ``ramp'' \\[-6pt]
3 & ``building/VBG permit/NN ./. '/'' Anyway/RB ,/, we/PRP should/MD have/VB
a/DT {\bf boat/NN} next/JJ summer/NN with/IN skiing/NN and/CC tubing/NN
paraphernalia/NNS ./.'' & ``permit'' \newline ``summer'' \\
\hline
\end{tabular}
} 
\end{center}
\caption{Examples of Step 2.1 in domain space for the $n$-gram {\em boat}. The
three tagged phrases generate five contextual patterns.}
\label{tab:dompats}
\end{table}

For Step 2.3, we set $n_c$ to 50,000. In Step 2.4, after we drop rows that are all zero,
we are left with $n_r$ equal to 114,297. After PPMI (which sets negative elements
to zero) we have 149,673,340 nonzero values, for a matrix density of 2.62\%.
Table~\ref{tab:domcols} shows the contextual patterns for the first five columns
and the last five columns (the columns are in order of their ranks in Step 2.2).
The {\em Count} column of the table gives the number of rows ($n$-grams) that generate
the pattern (that is, these are the counts mentioned in Step 2.2). The last patterns
all begin with {\em c} because they have the same counts and ties are broken by
alphabetical order. 

\begin{table}[h]
\begin{center}
\begin{minipage}{0.4\textwidth}
\scalebox{0.92}{
\begin{tabular}{clc}
\hline
Column & Pattern   & Count \\
\hline
1      & ``time''  & 91,483 \\
2      & ``part''  & 84,445 \\
3      & ``years'' & 84,417 \\
4      & ``way''   & 84,172 \\
5      & ``name''  & 81,960 \\
\hline
\end{tabular}
} 
\end{minipage}
\begin{minipage}{0.4\textwidth}
\scalebox{0.92}{
\begin{tabular}{clc}
\hline
Column & Pattern            & Count \\
\hline
49,996 & ``clu''            & 443 \\
49,997 & ``co-conspirator'' & 443 \\
49.998 & ``conciseness''    & 443 \\
49,999 & ``condyle''        & 443 \\
50,000 & ``conocer''        & 443 \\
\hline
\end{tabular}
} 
\end{minipage}
\end{center}
\caption{Contextual patterns for the first and last columns in domain space.
{\em CLU} is an abbreviation for Chartered Life Underwriter and other terms,
{\em condyle} is a round bump on a bone where it forms a joint with another
bone, and {\em conocer} is the Spanish verb {\em to know}, in the sense of
being acquainted with a person.}
\label{tab:domcols}
\end{table}

\subsection{Function Space}
\label{subsec:function}

The concept of function space is that the function or role of a word is
characterized by the syntactic context that relates it to the verbs
that occur near it. We use a more narrow window for function space than
domain space, based on the intuition that proximity to a verb is important
for determining the functional role of the given word. A distant verb
is less likely to characterize the function of the word. We generate relatively
complex patterns for function space, to try to capture the syntactic patterns
that connect the given word to the nearby verbs.

In Step 2.1, each tagged phrase generates up to six contextual patterns. For
a given tagged phrase, the first step is to cut the window down to 3 tokens before
the given \mbox{$n$-gram} and 3 tokens after it. If any of the remaining tokens to the left
of the \mbox{$n$-gram} are punctuation, the punctuation and everything to the left of the
punctuation is removed. If any of the remaining tokens to the right of the \mbox{$n$-gram}
are punctuation, the punctuation and everything to the right of the punctuation
is removed. Let's call the remaining tagged phrase a truncated tagged phrase.

Next we replace the given \mbox{$n$-gram} in the truncated tagged phrase with a generic
marker, X. We then simplify the part-of-speech tags by reducing them all to their first
character \cite{santorini90}. For example, all of the various verb tags (VB, VBD, VBG, VBN,
VBP, VBZ) are reduced to V. If the truncated tagged phrase contains no V tag, it generates
zero contextual patterns. If the phrase contains a V tag, then we generate two types of
contextual patterns, general patterns and specific patterns.

For the general patterns, the verbs (every token with a V tag) have their tags removed (naked
verbs) and all other tokens are reduced to naked tags (tags without words). For the specific
patterns, verbs, modals (tokens with M tags), prepositions (tokens with I tags), and {\em to}
(tokens with T tags) have their tags removed and all other tokens are reduced to naked tags.
(See Table~\ref{tab:funpats} for examples.)

For both general and specific patterns, to the left of X, we trim any leading naked tags.
To the right of X, we trim any trailing naked tags. A T tag can only be {\em to}, so we
replace any remaining naked T tags with {\em to}. A sequence of N tags (N N or N N N) is
likely a compound noun, so we reduce the sequence to a single N.

For a given truncated tagged phrase, we now have two patterns, one general pattern and one
specific pattern. If either of these patterns has tokens on both the left and right sides
of X, we make two more patterns by duplicating the X and then splitting the pattern at the
point between the two Xs. If one of the new patterns does not have a verb, we drop it. Thus
we may now have up to three specific patterns and three general patterns for the given
truncated tagged phrase. If the specific and general patterns are the same, only one of
them is generated.

Table~\ref{tab:funpats} shows some examples for the $n$-gram {\em boat}. Note
that every pattern must contain the generic marker, X, and at least one verb.

\begin{table}[h]
\begin{center}
\scalebox{0.92}{
\begin{tabular}{p{0.1in} p{3in} p{2in} p{0.6in}}
\hline
& Truncated tagged phrases & Patterns & Types \\
\hline
1 & ``the/DT canals/NNS by/IN {\bf boat/NN} and/CC wandering/VBG the/DT'' \newline
& ``X C wandering'' \newline ``by X C wandering'' & general \newline specific \\[-8pt]
2 & ``a/DT charter/NN fishing/VBG {\bf boat/NN} captain/NN named/VBN Jim/NNP'' \newline
\newline & ``fishing X N named'' \newline ``fishing X'' \newline ``X N named''
& general \newline general \newline general \\[-8pt]
3 & ``used/VBN from/IN a/DT {\bf boat/NN} and/CC lowered/VBD to/TO'' \newline
& ``used I D X C lowered'' \newline ``used I D X'' \newline ``X C lowered'' \newline 
``used from D X C lowered to'' \newline ``used from D X'' \newline ``X C lowered to''
& general \newline general \newline general \newline specific \newline specific
\newline specific \\
\hline
\end{tabular}
} 
\end{center}
\caption{Examples of Step 2.1 in function space for the $n$-gram {\em boat}. The
three truncated tagged phrases generate eleven contextual patterns.}
\label{tab:funpats}
\end{table}

For Step 2.3, we set $n_c$ to 50,000. In Step 2.4, after rows that are all zero are dropped,
$n_r$ is 114,101. After PPMI, there are 68,876,310 nonzero values, yielding a matrix
density of 1.21\%. Table~\ref{tab:funcols} shows the contextual patterns for the first
and the last five columns. The last patterns all begin with {\em s} because they have
the same counts and ties are broken by alphabetical order.

\begin{table}[h]
\begin{center}
\begin{minipage}{0.4\textwidth}
\scalebox{0.92}{
\begin{tabular}{clc}
\hline
Column & Pattern    & Count \\
\hline
1      & ``X is''   & 94,312 \\
2      & ``X N is'' & 82,171 \\
3      & ``is D X'' & 79,131 \\
4      & ``is X''   & 72,637 \\
5      & ``X was''  & 72,497 \\
\hline
\end{tabular}
} 
\end{minipage}
\begin{minipage}{0.4\textwidth}
\scalebox{0.92}{
\begin{tabular}{clc}
\hline
Column & Pattern               & Count \\
\hline
49,996 & ``since D X N was''   & 381 \\
49,997 & ``sinking I D X''     & 381 \\
49,998 & ``supplied with X''   & 381 \\
49,999 & ``supports D X N of'' & 381 \\
50,000 & ``suppressed I D X''  & 381 \\
\hline
\end{tabular}
} 
\end{minipage}
\end{center}
\caption{Contextual patterns for the first and last columns in function space.}
\label{tab:funcols}
\end{table}

The contextual patterns for function space are more complex than the patterns
for domain space. The motivation for this greater complexity is the observation
that mere proximity is not enough to determine functional roles, although it seems
sufficient for determining domains. For example, consider the verb {\em gives}.
If there is a word X that occurs near {\em gives}, X could be the subject, direct object,
or indirect object of the verb. To determine the functional role of X, we need to know which case
applies. The syntactic context that connects X to {\em gives} provides this information. The
contextual pattern ``X gives'' implies that X is the subject, ``gives X'' implies X is an object,
likely the direct object, and ``gives to X'' suggests that X is the indirect object.
Modals and prepositions supply further information about the functional role of X
in the context of a given verb. The verb {\em gives} appears in 43 different
contextual patterns (i.e., 43 of the 50,000 columns in function space correspond
to syntactic patterns that contain {\em gives}).

Many of the row vectors in the function space matrix correspond to verbs. It might seem
surprising that we can characterize the function of a verb by its syntactic relation to other
verbs, but consider an example, such as the verb {\em run}. The row vector for
{\em run} in the PPMI matrix for function space has 1,296 nonzero values; that is,
{\em run} is characterized by 1,296 different contextual patterns.

Note that appearing in a contextual pattern is different from having a nonzero value
for a contextual pattern. The character string for the word {\em run} appears in 62 different
contextual patterns, such as ``run out of X''. The row vector for the word {\em run}
has nonzero values for 1,296 contextual patterns (columns), such as ``had to X''.

\subsection{Mono Space}
\label{subsec:mono}

Mono space is simply the merger of domain space and function space.
For Step 2.3, we take the union of the 50,000 domain space columns
and the 50,000 function space columns, resulting in a total $n_c$ of
100,000 columns. In Step 2.4, we have a total $n_r$ of 114,297 rows.
The mono matrix after PPMI has 218,222,254 nonzero values, yielding a 
density of 1.91\%. The values in the mono frequency matrix $\mathbf{F}$
equal the corresponding values in the domain and function matrices.
Some of the rows in the mono space matrix do not have corresponding
rows in the function space matrix. For these rows, the corresponding
values are zeros (but there are nonzero elements in these rows, which
correspond to values in the domain matrix).

\subsection{Summary of the Spaces}
\label{subsec:summary}

Table~\ref{tab:spaces} summarizes the three matrices. In the following four sets of experiments,
we use the same three matrices (the domain, function, and mono matrices) in all cases; we do
not generate different matrices for each set of experiments. Three of the four sets of
experiments involve datasets that have been used in past by other researchers. We made no
special effort to ensure that the words in these three datasets have corresponding rows
in the three matrices. The intention is that these three matrices should be adequate to
handle most applications without any special customization.

\begin{table}[h]
\begin{center}
\scalebox{0.92}{
\begin{tabular}{lrrrr}
\hline
Space & Rows ($n_r$) & Columns ($n_c$) & Nonzeros (after PPMI) & Density (after PPMI) \\
\hline
domain   & 114,297 &  50,000  &         149,673,340 & 2.62\% \\
function & 114,101 &  50,000  &          68,876,310 & 1.21\% \\
mono     & 114,297 & 100,000  &         218,222,254 & 1.91\% \\
\hline
\end{tabular}
} 
\end{center}
\caption{Summary of the three spaces.}
\label{tab:spaces}
\end{table}

\subsection{Using the Spaces to Measure Similarity}
\label{subsec:using}

In the following experiments, we measure the similarity of two terms,
$a$ and $b$, by the cosine of the angle $\theta$ between their corresponding
row vectors, $\mathbf{a}$ and $\mathbf{b}$:

\begin{equation}
\label{eqn:cosine}
{\rm sim}(a,b) =
{\rm cos}(\mathbf{a},\mathbf{b}) =
\frac{{\mathbf{a}}}{{\left\| {\mathbf{a} } \right\| }}
\cdot \frac{{\mathbf{b}}}{{\left\| {\mathbf{b} } \right\| }}
\end{equation}

\noindent The cosine of the angle between two vectors is
the inner product of the vectors, after they have been normalized to unit length.
The cosine ranges from $-1$ when the vectors point in opposite directions
($\theta$ is 180 degrees) to $+1$ when they point in the same direction
($\theta$ is 0 degrees). When the vectors are orthogonal ($\theta$ is 90 degrees),
the cosine is zero. With raw frequency vectors, which necessarily cannot
have negative elements, the cosine cannot be negative, but weighting and smoothing
often introduce negative elements. PPMI weighting does not yield negative
elements, but truncated SVD can generate negative elements, even when the
input matrix has no negative values.

The semantic similarity of two terms is given by the cosine of the two corresponding rows in
$\mathbf{U}_k \mathbf{\Sigma}_k^p$ (see Section~\ref{subsec:building}). There are two parameters
in $\mathbf{U}_k \mathbf{\Sigma}_k^p$ that need to be set. The parameter $k$ controls the
number of latent factors and the parameter $p$ adjusts the weights of the factors, by raising
the corresponding singular values in $\mathbf{\Sigma}_k^p$ to the power $p$. The parameter
$k$ is well-known in the literature \cite{landauer07}, but $p$ is less familiar. The use of
$p$ was suggested by \citeA{caron01}. In the following experiments (Section~\ref{sec:experiments}),
we explore a range of values for $p$ and $k$.

Suppose we take a word $w$ and list all of the other words in descending order of their cosines
with $w$, using $\mathbf{U}_k \mathbf{\Sigma}_k^p$ to calculate the cosines. When $p$ is high,
as we go down the list, the cosines of the nearest neighbours of $w$ decrease slowly.
When $p$ is low, they decrease quickly. That is, a high $p$ results in a broad, fuzzy neighbourhood
and a low $p$ yields a sharp, crisp neighbourhood. The parameter $p$ controls the {\em sharpness}
of the similarity measure.

To reduce the running time of SVDLIBC, we limit the number of singular values
to 1500, which usually results in less than 1500 singular values. For example,
the SVD for domain space has 1477 singular values. As long as $k$ is not greater
than 1477, we can experiment with a range of $k$ values without rerunning SVDLIBC. We can
generate $\mathbf{U}_k \mathbf{\Sigma}_k^p$ from $\mathbf{U}_{1477} \mathbf{\Sigma}_{1477}^p$
by simply deleting the $1477 - k$ columns with the smallest singular values.

In the experiments, we vary $k$ from 100 to 1400 in increments of 100 (14 values for $k$) and we
vary $p$ from $-1$ to $+1$ in increments of 0.1 (21 values for $p$). When $p$ is $-1$, we give more
weight to the factors with smaller singular values; when $p$ is $+1$, the factors with larger
singular values have more weight. \citeA{caron01} observes that most researchers use either
$p = 0$ or $p = 1$; that is, they use either $\mathbf{U}_k$ or $\mathbf{U}_k \mathbf{\Sigma}_k$.

Let ${\rm sim_f}(a,b) \in \Re$ be function similarity as measured by
the cosine of vectors $\mathbf{a}$ and $\mathbf{b}$ in function space. Let
${\rm sim_d}(a,b) \in \Re$ be domain similarity as measured by
the cosine of vectors $\mathbf{a}$ and $\mathbf{b}$ in domain space.
When a similarity measure combines both ${\rm sim_d}(a,b)$ and ${\rm sim_f}(a,b)$,
there are four parameters to tune, $k_{\rm d}$ and $p_{\rm d}$ for domain space and
$k_{\rm f}$ and $p_{\rm f}$ for function space.

For one space, it is feasible for us to explore all $14 \times 21 = 294$ combinations
of parameter values, but two spaces have $294 \times 294 = 86,436$ combinations of values.
To make the search tractable, we initialize the parameters to the middle of their ranges
($k_{\rm f} = k_{\rm d} = 700$ and $p_{\rm f} = p_{\rm d} = 0$) and then we alternate
between tuning ${\rm sim_d}(a,b)$ (i.e., $k_{\rm d}$ and $p_{\rm d}$) while holding
${\rm sim_f}(a,b)$ (i.e., $k_{\rm f}$ and $p_{\rm f}$) fixed and tuning ${\rm sim_f}(a,b)$
while holding ${\rm sim_d}(a,b)$ fixed. We stop the search when there is no improvement in
performance on the training data. In almost all cases, a local optimum is found in one pass;
that is, after we have tuned the parameters once, there is no improvement when we try to
tune them a second time. Thus we typically evaluate $294 \times 3 = 882$ parameter values
($\times 3$ because we tune one similarity, tune the other, and then try the first again to
see if further improvement is possible).\footnote{We use Perl Data Language (PDL) for searching
for parameters, calculating cosines, and other operations on vectors and matrices. See
http://pdl.perl.org/.}

We could use a standard numerical optimization algorithm to tune the four
parameters, but the algorithm we use here takes advantage of background knowledge about
the optimization task. We know that small variations in the parameters make small changes in
performance, so there is no need to make a very fine-grained search, and we know that
${\rm sim_d}(a,b)$ and ${\rm sim_f}(a,b)$ are relatively independent, so we can optimize
them separately.

The rows in the matrices are based on terms in the WordNet {\em index.sense} file.
In this file, all nouns are in their singular forms and all verbs are in their stem forms.
To calculate ${\rm sim}(a,b)$, we first look for exact matches for $a$ and $b$ in
the terms that correspond to the rows of the given matrix (domain, function, or mono).
If an exact match is found, then we use the corresponding row vector in the matrix.
Otherwise, we look for alternate forms of the terms, using the {\em validForms}
function in the WordNet::QueryData Perl interface to WordNet.\footnote{WordNet::QueryData
is available at http://search.cpan.org/dist/WordNet-QueryData/.} This automatically
converts plural nouns to their singular forms and verbs to their stem forms.
If none of the alternate forms is an exact match for a row in the matrix, we map the term to a
zero vector of length $k$.

\subsection{Composing Similarities}
\label{subsec:composing}

Our approach to semantic relations and compositions is to combine the two similarities,
${\rm sim_d}(a,b)$ and ${\rm sim_f}(a,b)$, in various ways, depending on the task at hand
or the syntax of the phrase at hand. In general, we want the combined similarity to be
high when the component similarities are high, and we want the values of the
component similarities to be balanced. To achieve balance, we use the geometric mean
to combine similarities, instead of the arithmetic mean. The geometric mean is not
suitable for negative numbers, and the cosine can be negative in some cases; hence
we define the geometric mean as zero if any of the component similarities are negative:

\begin{equation}
\label{eqn:geo} {\rm geo}(x_1, x_2, \dots, x_n) =
\left\{
\begin{array}{rl}
(x_1 x_2 \dots x_n)^{1/n}
& \mbox{if $x_i > 0$ for all $i = 1, \dots, n$} \\
0 & \mbox{otherwise}
\end{array}
\right.
\end{equation}

\subsection{Element-wise Multiplication}
\label{subsec:multiplication}

One of the most successful approaches to composition, so far, has been
element-wise multiplication, $\mathbf{c} = \mathbf{a} \odot \mathbf{b}$, where
$c_i = a_i \, \cdot \, b_i$ \cite{mitchell08,mitchell10}. This approach only
makes sense when the elements in the vectors are not negative. When the elements
in $\mathbf{a}$ and $\mathbf{b}$ are positive, relatively large values of
$a_i$ and $b_i$ reinforce each other, resulting in a large value for $c_i$.
This makes intuitive sense. But when $a_i$ and $b_i$ are both highly
negative, $c_i$ will be highly positive, although intuition says $c_i$ should
be highly negative. \citeA{mitchell08,mitchell10} designed their word--context
matrices to ensure that the vectors had no negative elements.

The values in the matrix $\mathbf{U}_k \mathbf{\Sigma}_k^p$ are typically about
half positive and half negative. We use element-wise multiplication as a baseline
in some of the following experiments. For a fair baseline, we cannot simply apply
element-wise multiplication to row vectors in $\mathbf{U}_k \mathbf{\Sigma}_k^p$.
One solution would be to use the PPMI matrix, $\mathbf{X}$, which has no
negative elements, but this would not allow element-wise multiplication to
take advantage of the smoothing effect of SVD. Our solution is to use row vectors
from ${\bf \hat X} = \mathbf{U}_k \mathbf{\Sigma}_k \mathbf{V}_k^\mathsf{T}$.
Although the PPMI matrix, $\mathbf{X}$, is sparse (see Table~\ref{tab:spaces}),
${\bf \hat X}$ and $\mathbf{U}_k \mathbf{\Sigma}_k^p$ have a density of 100\%.

Let $\mathbf{a}'$ and $\mathbf{b}'$ be the vectors in ${\bf \hat X}$ that
correspond to the terms $a$ and $b$. These row vectors benefit from smoothing
due to truncated SVD, but their elements are almost all positive. If there are any
negative elements, we set them to zero. Let $\mathbf{c}' = \mathbf{a}' \odot \mathbf{b}'$.
After we apply element-wise multiplication to the vectors, we then multiply by
$\mathbf{V}_k \mathbf{\Sigma}_k^{p-1}$, so that the resulting vector
$\mathbf{c} = \mathbf{c}' \mathbf{V}_k \mathbf{\Sigma}_k^{p-1}$ can be compared
with other row vectors in the matrix $\mathbf{U}_k \mathbf{\Sigma}_k^p$:

\begin{align}
{\bf \hat X} (\mathbf{V}_k \mathbf{\Sigma}_k^{p-1})
& =(\mathbf{U}_k \mathbf{\Sigma}_k \mathbf{V}_k^\mathsf{T}) (\mathbf{V}_k \mathbf{\Sigma}_k^{p-1}) \\
& = \mathbf{U}_k \mathbf{\Sigma}_k \mathbf{V}_k^\mathsf{T} \mathbf{V}_k \mathbf{\Sigma}_k^{p-1} \\
& = \mathbf{U}_k \mathbf{\Sigma}_k  \mathbf{\Sigma}_k^{p-1} \\
& = \mathbf{U}_k \mathbf{\Sigma}_k^p  
\end{align}

\noindent Note that, since $\mathbf{V}_k$ is column orthonormal, $\mathbf{V}_k^\mathsf{T} \mathbf{V}_k$
equals $\mathbf{I}_k$, the $k \times k$ identity matrix.

Similarly, if $\mathbf{a}$ is a row vector in $\mathbf{U}_k \mathbf{\Sigma}_k^p$,
we can find its counterpart $\mathbf{a}'$ in ${\bf \hat X}$ by multiplying $\mathbf{a}$ with
$\mathbf{\Sigma}_k^{1-p} \mathbf{V}_k^\mathsf{T}$:

\begin{align}
(\mathbf{U}_k \mathbf{\Sigma}_k^p) (\mathbf{\Sigma}_k^{1-p} \mathbf{V}_k^\mathsf{T})
& = \mathbf{U}_k \mathbf{\Sigma}_k^p \mathbf{\Sigma}_k^{1-p} \mathbf{V}_k^\mathsf{T} \\
& = \mathbf{U}_k \mathbf{\Sigma}_k \mathbf{V}_k^\mathsf{T} \\
& = {\bf \hat X}
\end{align}

Let ${\rm nn}(\mathbf{x})$ ({\em nn} for {\em nonnegative}) be a function that
converts negative elements in a vector $\mathbf{x}$ to zero:

\begin{align}
{\rm nn}(\langle x_1, \ldots, x_n \rangle) & = \langle y_1, \ldots, y_n \rangle \\
y_i & =
\left\{
\begin{array}{rl}
x_i
& \mbox{if $x_i > 0$} \\
0 & \mbox{otherwise}
\end{array}
\right.
\end{align}

\noindent Our version of element-wise multiplication may be expressed
as follows:

\begin{equation}
\label{eqn:mult}
\mathbf{c} =
({\rm nn}(\mathbf{a} \mathbf{\Sigma}_k^{1-p} \mathbf{V}_k^\mathsf{T}) \odot
{\rm nn}(\mathbf{b} \mathbf{\Sigma}_k^{1-p} \mathbf{V}_k^\mathsf{T})) \,
\mathbf{V}_k \mathbf{\Sigma}_k^{p-1}
\end{equation}

Another way to deal with element-wise multiplication would be to use
nonnegative matrix factorization (NMF) \cite{lee99} instead of SVD.
We have not yet found an implementation of NMF that scales to the matrix sizes that
we have here (Table~\ref{tab:spaces}). In our past experiments with smaller
matrices, SVD and NMF have similar performance.

\section{Experiments with Varieties of Similarities}
\label{sec:experiments}

This section presents four sets of experiments. The first set of experiments
presents a dual-space model of semantic relations and evaluates the
model with multiple choice analogy questions from the SAT. The second
set presents a model of semantic composition and evaluates it with
multiple choice questions that are constructed from WordNet.
The third set applies a dual-space model to the phrase similarity
dataset of \citeA{mitchell10}. The final set uses three classes
of word pairs from \citeA{chiarello90} to test a hypothesis about
the dual-space model, that domain space and function space capture
the intuitive concepts of {\em association} and {\em similarity}.

\subsection{Similarity of Relations}
\label{subsec:sat-exper}

Here we evaluate the dual-space model applied to the task of measuring
the similarity of semantic relations. We use a set of 374 multiple-choice
analogy questions from the SAT college entrance exam \cite{turney06b}.
Table~\ref{tab:sat} gives an example of one of the questions. The task
is to select the choice word pair that is most analogous (most relationally
similar) to the stem word pair.

\begin{table}[h]
\begin{center}
\scalebox{0.92}{
\begin{tabular}{lll}
\hline
Stem:      &       & lull:trust \\
\hline
Choices:   & (1)   & balk:fortitude \\
           & (2)   & betray:loyalty \\
           & (3)   & cajole:compliance \\
           & (4)   & hinder:destination \\
           & (5)   & soothe:passion \\
\hline
Solution:  & (3)   & cajole:compliance \\
\hline
\end{tabular}
} 
\end{center}
\caption {An example of a question from the 374 SAT analogy questions. Lulling
a person into trust is analogous to cajoling a person into compliance.}
\label{tab:sat}
\end{table}

Let $a\!:\!b$ represent the stem pair (e.g., {\em lull}$\,:${\em trust}). We answer
the SAT questions by selecting the choice pair $c\!:\!d$ that maximizes the
relational similarity, ${\rm sim_r}(a\!:\!b, c\!:\!d)$, defined as follows:

\begin{align}
\label{eqn:sat1} {\rm sim_1}(a\!:\!b, c\!:\!d) & = {\rm geo}({\rm sim_f}(a,c), {\rm sim_f}(b,d)) \\
\label{eqn:sat2} {\rm sim_2}(a\!:\!b, c\!:\!d) & = {\rm geo}({\rm sim_d}(a,b), {\rm sim_d}(c,d)) \\
\label{eqn:sat3} {\rm sim_3}(a\!:\!b, c\!:\!d) & = {\rm geo}({\rm sim_d}(a,d), {\rm sim_d}(c,b)) \\
\label{eqn:sat4} {\rm sim_r}(a\!:\!b, c\!:\!d) & =
\left\{
\begin{array}{rl}
{\rm sim_1}(a\!:\!b, c\!:\!d) & \mbox{if ${\rm sim_2}(a\!:\!b, c\!:\!d) \ge {\rm sim_3}(a\!:\!b, c\!:\!d)$} \\
0 & \mbox{otherwise}
\end{array}
\right.
\end{align}

\noindent The intent of ${\rm sim_1}$ is to measure the function similarity across the
two pairs. The domain similarity {\em inside} the two pairs is measured by ${\rm sim_2}$,
whereas the domain similarity {\em across} the two pairs is given by ${\rm sim_3}$.
The relational similarity, ${\rm sim_r}$, is simply the function similarity, ${\rm sim_1}$,
subject to the constraint that the domain similarity inside pairs, ${\rm sim_2}$, must not be
less than the domain similarity across pairs, ${\rm sim_3}$.

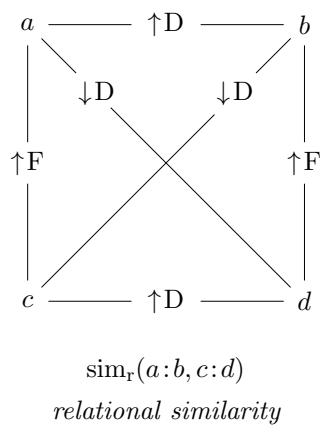
\begin{figure}[h]
\begin{center}
\scalebox{0.92}{
\begin{tikzpicture}
\node at (0,5) {$a$};
\node at (4,5) {$b$};
\node at (0,1) {$c$};
\node at (4,1) {$d$};
\node at (2,5) {$\uparrow \! {\rm D}$};
\node at (2,1) {$\uparrow \! {\rm D}$};
\node at (0,3) {$\uparrow \! {\rm F}$};
\node at (4,3) {$\uparrow \! {\rm F}$};
\node at (1,4) {$\downarrow \! {\rm D}$};
\node at (3,4) {$\downarrow \! {\rm D}$};
\node at (2,0) {${\rm sim_r}(a\!:\!b, c\!:\!d)$};
\node at (2.0,-0.6) {\em relational similarity};
\draw (0.3,5) -- (1.5,5);       
\draw (2.5,5) -- (3.7,5);       
\draw (0.3,1) -- (1.5,1);       
\draw (2.5,1) -- (3.7,1);       
\draw (0,1.3) -- (0,2.7);       
\draw (0,3.3) -- (0,4.7);       
\draw (4,1.3) -- (4,2.7);       
\draw (4,3.3) -- (4,4.7);       
\draw (0.2,1.2) -- (2.7,3.7);   
\draw (3.3,4.3) -- (3.8,4.8);   
\draw (0.2,4.8) -- (0.7,4.3);   
\draw (1.2,3.8) -- (3.8,1.2);   
\end{tikzpicture}
} 
\end{center}
\caption{A diagram of the reasoning behind Equations \ref{eqn:sat1} to \ref{eqn:sat4}.
$\uparrow \! {\rm F}$ represents high function similarity, $\uparrow \! {\rm D}$ means
high domain similarity, and $\downarrow \! {\rm D}$ indicates low domain similarity.}
\label{fig:simr}
\end{figure}

Figure~\ref{fig:simr} conveys the main ideas behind Equations \ref{eqn:sat1} to \ref{eqn:sat4}.
We want high function similarities (indicated by $\uparrow \! {\rm F}$) for $a\!:\!c$ and $b\!:\!d$,
as measured by ${\rm sim_1}$. We also prefer relatively high domain similarities
(marked with $\uparrow \! {\rm D}$) for $a\!:\!b$ and $c\!:\!d$ (measured by ${\rm sim_2}$),
in contrast to relatively low domain similarities ($\downarrow \! {\rm D}$) for
$a\!:\!d$ and $c\!:\!b$ (as given by ${\rm sim_3}$).\footnote{We recently came across this
same rectangular structure in \citeS{lepage96} paper on morphological analogy (see their
Figure 1). Although our algorithm and our task differ considerably from the algorithm and task
of \citeA{lepage96}, we have independently discovered the same underlying structure in
analogical reasoning.}

Using the example in Table~\ref{tab:sat}, we see that lulling a person into trust is analogous
to cajoling a person into compliance, since the functional role of {\em lull} is
similar to the functional role of {\em cajole} (both involve manipulating a person)
and the functional role of {\em trust} is similar to the functional role of {\em compliance}
(both are states that a person can be in). This is captured by ${\rm sim_1}$. The constraint
${\rm sim_2}(a\!:\!b, c\!:\!d) \ge {\rm sim_3}(a\!:\!b, c\!:\!d)$ implies that the domain
similarities of {\em lull}$\,:${\em trust} (the domain of confidence and loyalty) and
{\em cajole}$\,:${\em compliance} (the domain of obedience and conformity)
should be greater than or equal to the domain similarities of {\em lull}$\,:${\em compliance}
and {\em cajole}$\,:${\em trust}.

Analogy is a way of mapping knowledge from a source domain to a target domain \cite{gentner83}.
If $a$ in the source domain is mapped to $c$ in the target domain, then $a$ should
play the same role in the source domain as $c$ plays in the target domain. This is the
theory behind ${\rm sim_1}$. If $a$ and $b$ are in the source domain and $c$ and $d$ are
in the target domain, then the internal domain similarity of $a$ and $b$ and the
internal domain similarity of $c$ and $d$ should not be less than the cross-domain
similarities. This motivates the constraint ${\rm sim_2} \ge {\rm sim_3}$.
Our definition is a natural expression of \citeS{gentner83} theory of analogy.

Recall the four equations that we introduced in Section~\ref{subsec:relations}.
We repeat these equations here for convenience:

{\allowdisplaybreaks 
\begin{align}
\label{eqn:copysimr1} {\rm sim_r}(a\!:\!b, c\!:\!d) & = {\rm sim_r}(b\!:\!a, d\!:\!c) \\
\label{eqn:copysimr2} {\rm sim_r}(a\!:\!b, c\!:\!d) & = {\rm sim_r}(c\!:\!d, a\!:\!b) \\
\label{eqn:copysimr3} {\rm sim_r}(a\!:\!b, c\!:\!d) & \neq {\rm sim_r}(a\!:\!b, d\!:\!c) \\
\label{eqn:copysimr4} {\rm sim_r}(a\!:\!b, c\!:\!d) & \neq {\rm sim_r}(a\!:\!d, c\!:\!b)
\end{align}
} 

\noindent Inspection will show that the definition of relational similarity in Equation~\ref{eqn:sat4}
satisfies the requirements of Equations \ref{eqn:copysimr1}, \ref{eqn:copysimr2}, \ref{eqn:copysimr3},
and \ref{eqn:copysimr4}. This can be understood by considering Figure~\ref{fig:simr}.
Equation~\ref{eqn:copysimr1} tells us that we can rotate Figure~\ref{fig:simr} about
its vertical axis without altering the network of similarities, due to the symmetry of the
figure. Equation~\ref{eqn:copysimr2} tells us that we can rotate Figure~\ref{fig:simr} about
its horizontal axis without altering the network of similarities.

On the other hand, we cannot swap $c$ and $d$ while holding $a$ and $b$ fixed, because this
would change both the $\uparrow \! {\rm F}$ and $\downarrow \! {\rm D}$ links (although it
would not change the $\uparrow \! {\rm D}$ links). In other words, ${\rm sim_1}$ and
${\rm sim_3}$ would be changed, although ${\rm sim_2}$ would not be affected. Therefore
Equation~\ref{eqn:copysimr3} is satisfied.

Also, we cannot swap $b$ and $d$ while holding $a$ and $c$ fixed, because this would change
the $\uparrow \! {\rm D}$ and $\downarrow \! {\rm D}$ links (although it would not change
the $\uparrow \! {\rm F}$ links). In other words, ${\rm sim_2}$ and ${\rm sim_3}$ would be
changed, although ${\rm sim_1}$ would not be affected. Therefore Equation~\ref{eqn:copysimr4}
is satisfied. We can see that ${\rm sim_1}$ by itself would violate Equation~\ref{eqn:copysimr4},
due to the symmetry of cosines, ${\rm sim_f}(b,d) = {\rm sim_f}(d,b)$. The constraint
${\rm sim_2}(a\!:\!b, c\!:\!d) \ge {\rm sim_3}(a\!:\!b, c\!:\!d)$ breaks this symmetry.

Another way to break the symmetry, so that Equation~\ref{eqn:copysimr4} is
satisfied, would be to use a similarity measure that is inherently asymmetric, such as skew
divergence. In Equation~\ref{eqn:sat4}, the symmetry is broken in a natural way by considering how
domain and function similarity apply to analogies, so there is no need to introduce an inherently
asymmetric measure. Also, note that the symmetries of Equations \ref{eqn:copysimr1} and
\ref{eqn:copysimr2} are desirable; we do not wish to break these symmetries.

It would have been reasonable to include ${\rm sim_d}(a,c)$ and ${\rm sim_d}(b,d)$
in ${\rm sim_3}$, but we decided to leave them out. It seems to us
that the function similarities ${\rm sim_f}(a,c)$ and ${\rm sim_f}(b,d)$,
which should have high values in a good analogy, might
cause ${\rm sim_d}(a,c)$ and ${\rm sim_d}(b,d)$ to be relatively high, even though
they cross domains. If people observe a certain kind of abstract function similarity frequently,
that function similarity might become a popular topic for discussion, which could result in
a high domain similarity.

For example, {\em carpenter}$\,:${\em wood} is analogous to
{\em mason}$\,:${\em stone}. The domain of {\em carpenter}$\,:${\em wood} is carpentry
and the domain of {\em mason}$\,:${\em stone} is masonry. The functional role of {\em carpenter}
is similar to the functional role of {\em mason}, in that both are artisans. Although
{\em carpenter} and {\em mason} belong to different domains, their high degree
of abstract function similarity may result in discussions that mention them
together, such as discussions about specialized trades, skilled manual labour,
the construction industry, and workplace injuries. In other words, high function
similarity between two words may cause a rise in their domain similarity. Therefore
we did not include ${\rm sim_d}(a,c)$ and ${\rm sim_d}(b,d)$ in ${\rm sim_3}$.

When all five choices for a SAT question have a relational similarity of zero, we skip the
question. We use ten-fold cross-validation to set the parameters for the SAT questions.
The same parameter values are selected in nine of the ten folds, $k_{\rm d} = 800$,
$p_{\rm d} = -0.1$, $k_{\rm f} = 300$, and $p_{\rm f} = 0.5$. After the parameters are
determined, all 374 SAT questions can be answered in a few seconds. Equation~\ref{eqn:sat4}
correctly answers 191 of the questions, skips 2 questions, and incorrectly answers 181
questions, achieving an accuracy of 51.1\%.

\subsubsection{Comparison with Past Work}

For comparison, the average score for senior highschool students applying to US
universities is 57.0\%. The ACL Wiki lists many past results with the 374 SAT
questions.\footnote{See http://aclweb.org/aclwiki/index.php?title=SAT\_Analogy\_Questions.}
Table~\ref{tab:topsat} shows the top ten results at the time of writing. In this table,
{\em dual-space} refers to the dual-space model using Equation~\ref{eqn:sat4}.
Four of the past results achieved an accuracy of 51.1\% or higher. All four used holistic
approaches and hence are not able to address the issue of linguistic creativity. The best
previous algorithm attains an accuracy of 56.1\% (210 correct, 4 skipped, 160 incorrect)
\cite{turney06b}. The difference between 51.1\% and 56.1\% is not statistically significant
at the 95\% confidence level, according to Fisher's Exact Test.

\begin{table}[h]
\begin{center}
\scalebox{0.92}{
\begin{tabular}{llcc}
\hline
Algorithm         & Reference                                   & Accuracy & 95\% confidence \\
\hline
LSA+Predication   & \citeA{mangalath04}                         & 42.0     & 37.2--47.4 \\
KNOW-BEST         & \citeA{veale04}                             & 43.0     & 38.0--48.2 \\
k-means           & Bi{\c{c}}ici and Yuret~\citeyear{bicici06}  & 44.0     & 39.0--49.3 \\
BagPack           & \citeA{herdagdelen09}                       & 44.1     & 39.0--49.3 \\
VSM               & \citeA{turney05a}                           & 47.1     & 42.2--52.5 \\
Dual-Space        &                                             & 51.1     & 46.1--56.5 \\
BMI               & \citeA{bollegala09}                         & 51.1     & 46.1--56.5 \\
PairClass         & \citeA{turney08a}                           & 52.1     & 46.9--57.3 \\
PERT              & \citeA{turney06a}                           & 53.5     & 48.5--58.9 \\
LRA               & \citeA{turney06b}                           & 56.1     & 51.0--61.2 \\
Human             & Average US college applicant                & 57.0     & 52.0--62.3 \\
\hline
\end{tabular}
} 
\end{center}
\caption {The top ten results with the 374 SAT questions, from the ACL Wiki. The 95\%
confidence intervals are calculated using the Binomial Exact Test.}
\label{tab:topsat}
\end{table}

The majority of the algorithms in Table~\ref{tab:topsat} are unsupervised, but Dual-Space,
PairClass \cite{turney08a}, and BagPack \cite{herdagdelen09} use limited supervision.
PairClass and BagPack answer a given SAT question by learning a binary classification model
that is specific to the given question. The training set for a given question consists of
one positive training example, the stem pair for the question, and ten randomly
selected pairs as (assumed) negative training examples. The induced binary classifier is used
to assign probabilities to the five choices and the most probable choice is the guess.
Dual-Space uses the training set only to tune four numerical parameters. These
three algorithms are best described as {\em weakly} supervised.

\subsubsection{Sensitivity to Parameters}

To see how sensitive the dual-space model is to the values of the parameters,
we perform two exhaustive grid searches, one with a coarse, wide grid and another
with a fine, narrow grid. For each point in the grids, we evaluate the dual-space
model using the whole set of 374 SAT questions. The narrow grid search is
centred on the parameter values that were selected in nine of the ten folds in the
previous experiment, $k_{\rm d} = 800$, $p_{\rm d} = -0.1$, $k_{\rm f} = 300$, and
$p_{\rm f} = 0.5$. Both searches evaluate 5 values for each parameter, yielding
a total of $5^4 = 625$ parameter settings. Table~\ref{tab:grid} shows the values that
were explored in the two grid searches and Table~\ref{tab:sensitivity} presents the
minimum, maximum, average, and standard deviation of the accuracy for the two searches.

\begin{table}[h]
\begin{center}
\scalebox{0.92}{
\begin{tabular}{lcrrrrr}
\hline
Grid          & Parameter   & \multicolumn{5}{c}{Values} \\
\hline
Coarse        & $k_{\rm d}$ & 100   & 425   & 750   & 1075  & 1400 \\
              & $p_{\rm d}$ & -1.0  & -0.5  & 0.0   & 0.5   & 1.0 \\
              & $k_{\rm f}$ & 100   & 425   & 750   & 1075  & 1400 \\
              & $p_{\rm f}$ & -1.0  & -0.5  & 0.0   & 0.5   & 1.0 \\
\hline
Fine          & $k_{\rm d}$ & 600   & 700   & 800   & 900   & 1000 \\
              & $p_{\rm d}$ & -0.3  & -0.2  & -0.1  & 0.0   & 0.1 \\
              & $k_{\rm f}$ & 100   & 200   & 300   & 400   & 500 \\
              & $p_{\rm f}$ & 0.3   & 0.4   & 0.5   & 0.6   & 0.7 \\
\hline
\end{tabular}
} 
\end{center}
\caption {The range of parameter values for the two grid searches.}
\label{tab:grid}
\end{table}

\begin{table}[h]
\begin{center}
\scalebox{0.92}{
\begin{tabular}{lcccc}
\hline
           & \multicolumn{4}{c}{Accuracy} \\
\cline{2-5}
Grid       & Minimum     & Maximum    & Average   & Standard deviation \\
\hline
Coarse     & 31.0        & 48.7       & 40.7      & 4.1 \\
Fine       & 42.5        & 51.6       & 47.3      & 2.0 \\
\hline
\end{tabular}
} 
\end{center}
\caption {The sensitivity of the dual-space model to the parameter settings.}
\label{tab:sensitivity}
\end{table}

The accuracy attained by the heuristic search (described in Section~\ref{subsec:using})
with ten-fold cross-validation, 51.1\% (Table~\ref{tab:topsat}), is near the best accuracy
of the fine grid search using the whole set of 374 SAT questions, 51.6\% (Table~\ref{tab:sensitivity}).
This is evidence that the heuristic search is effective. Accuracy with
the coarse search varies from 31.0\% to 48.7\%, which demonstrates the importance
of tuning the parameters. On the other hand, accuracy with the fine search
spans a narrower range and has a lower standard deviation, which suggests
that the dual-space model is not overly sensitive to relatively small
variations in the parameter values; that is, the parameters are reasonably stable.
(That nine of the ten folds in cross-validation select the same parameters
is further evidence of stability.)

\subsubsection{Parts of Speech}

Since domain space is based on nouns and function space is based on verbs, it
is interesting to know how the performance of the dual-space model varies with different
parts of speech. To answer this, we manually labeled all 374 SAT questions with
part-of-speech labels. The labels for a single pair can be ambiguous, but
the labels become unambiguous in the context of the whole question. For example,
{\em lull}$\,:${\em trust} could be {\em noun}$\,:${\em verb}, but in the context
of Table~\ref{tab:sat}, it must be {\em verb}$\,:${\em noun}.

Table~\ref{tab:pos} splits out the results for the various parts of speech.
None of the differences in this table are statistically significant at the 95\%
confidence level, according to Fisher's Exact Test. A larger and more varied set
of questions will be needed to determine how part of speech affects the dual-space
model.

\begin{table}[h]
\begin{center}
\scalebox{0.92}{
\begin{tabular}{lrrrrr}
\hline
Parts of speech                   &  Right  & Accuracy &  Wrong &  Skipped &  Total \\
\hline
noun:noun                         &     97  &   50.8   &   93   &    1     &  191 \\
noun:adjective or adjective:noun  &     35  &   53.0   &   31   &    0     &   66 \\
noun:verb or verb:noun            &     27  &   49.1   &   28   &    0     &   55 \\
adjective:adjective               &      9  &   37.5   &   15   &    0     &   24 \\
verb:adjective or adjective:verb  &     12  &   60.0   &    7   &    1     &   20 \\
verb:verb                         &     11  &   64.7   &    6   &    0     &   17 \\
verb:adverb or adverb:verb        &      0  &    0.0   &    1   &    0     &    1 \\
\hline
all                               &    191  &   51.1   &  181   &    2     &  374 \\
\hline
\end{tabular}
} 
\end{center}
\caption {Performance of the dual-space model with various parts of speech.}
\label{tab:pos}
\end{table}

\subsubsection{Order Sensitivity}

It seems that function space is doing most of the work in Equation~\ref{eqn:sat4}.
If we use ${\rm sim_1}$ alone, dropping the constraint that ${\rm sim_2} \ge {\rm sim_3}$,
then accuracy drops from 51.1\% to 50.8\%. This drop is not statistically significant. We hypothesize
that the small drop is due to the design of the SAT test, which is primarily intended
to test a student's understanding of functional roles, not domains.

To verify this hypothesis, we reformulated the SAT questions so that they would test both
function and domain comprehension. The method is to first expand each choice pair $c\!:\!d$
by including the stem pair $a\!:\!b$, resulting in the full explicit analogy $a\!:\!b\!::\!c\!:\!d$.
For each expanded choice, $a\!:\!b\!::\!c\!:\!d$, we then generate another choice,
$a\!:\!d\!::\!c\!:\!b$. Table~\ref{tab:expanded} shows the reformulation of
Table~\ref{tab:sat}. Due to symmetry, ${\rm sim_1}$ must assign the same similarity
to both $a\!:\!b\!::\!c\!:\!d$ and $a\!:\!d\!::\!c\!:\!b$. This new ten-choice
test evaluates both function and domain similarities.

\begin{table}[h]
\begin{center}
\scalebox{0.92}{
\begin{tabular}{lll}
\hline
Choices:  & (1)  & lull:trust::balk:fortitude \\
          & (2)  & lull:fortitude::balk:trust \\
          & (3)  & lull:loyalty::betray:trust \\
          & (4)  & lull:trust::betray:loyalty \\
          & (5)  & lull:compliance::cajole:trust \\
          & (6)  & lull:trust::cajole:compliance \\
          & (7)  & lull:destination::hinder:trust \\
          & (8)  & lull:trust::hinder:destination \\
          & (9)  & lull:trust::soothe:passion \\
          & (10) & lull:passion::soothe:trust \\
\hline
Solution: & (6)  & lull:trust::cajole:compliance \\
\hline
\end{tabular}
} 
\end{center}
\caption {An expanded SAT question, designed to test both function and domain
comprehension. Choices (5) and (6) have the same similarity according to ${\rm sim_1}$.}
\label{tab:expanded}
\end{table}

The task with the expanded ten-choice SAT questions is the same
as with the original five-choice questions, to select the best analogy.
The solution in Table~\ref{tab:expanded} is the same as the solution in
Table~\ref{tab:sat}, except that the stem pair is explicit in Table~\ref{tab:expanded}.
The only signficant change is that five new distractors have been added to the
choices. We answer the ten-choice questions by selecting the choice $a\!:\!b\!::\!c\!:\!d$
that maximizes ${\rm sim_r}(a\!:\!b, c\!:\!d)$.

On the ten-choice reformulated SAT test, ${\rm sim_r}$ (Equation~\ref{eqn:sat4}) attains an
accuracy of 47.9\%, whereas ${\rm sim_1}$ alone (Equation~\ref{eqn:sat1}) 
only achieves 27.5\%. The difference is statistically significant at the 95\% confidence
level, according to Fisher's Exact Test. This more stringent test supports the claim
that function similarity is insufficient by itself.

As a further test of the value of two separate spaces, we use a single space
for both ${\rm sim_d}$ and ${\rm sim_f}$ in Equation~\ref{eqn:sat4}. The model
still has four parameters it can tune, $k_{\rm d}$, $p_{\rm d}$, $k_{\rm f}$, and $p_{\rm f}$,
but the same matrix is used for both similarities. The best result is an
accuracy of 40.4\% on the ten-question reformulated SAT test, using function space
for both ${\rm sim_d}$ and ${\rm sim_f}$. This is significantly below the 47.9\% accuracy of the
dual-space model when ${\rm sim_d}$ is based on domain space and ${\rm sim_f}$ is
based on function space (95\% confidence level, Fisher's Exact Test).

Table~\ref{tab:satsum} summarizes the results. In the cases where the matrix for
${\rm sim_d}$ is {\em not used}, the model is based on ${\rm sim_1}$ alone
(Equation~\ref{eqn:sat1}). In all other cases, the model is based on ${\rm sim_r}$
(Equation~\ref{eqn:sat4}). For both the five-choice and ten-choice SAT questions,
the original dual-space model is more accurate than any of the modified models.
The {\em Significant} column indicates whether the accuracy of a modified model
is significantly less than the original dual-space model (95\% confidence level,
Fisher's Exact Test). The more difficult ten-choice questions clearly show the value
of two distinct spaces.

\begin{table}[h]
\begin{center}
\scalebox{0.92}{
\begin{tabular}{lcclll}
\hline
Algorithm            & Accuracy & Significant & Questions      & Matrix for ${\rm sim_d}$ & Matrix for ${\rm sim_f}$ \\
\hline
dual-space           & 51.1     &             & five-choice    & domain space             & function space \\
modified dual-space  & 47.3     & no          & five-choice    & function space           & function space \\
modified dual-space  & 43.6     & yes         & five-choice    & mono space               & mono space \\
modified dual-space  & 37.7     & yes         & five-choice    & domain space             & domain space \\
modified dual-space  & 50.8     & no          & five-choice    & not used                 & function space \\
modified dual-space  & 41.7     & yes         & five-choice    & not used                 & mono space \\
modified dual-space  & 35.8     & yes         & five-choice    & not used                 & domain space \\
\hline
dual-space           & 47.9     &             & ten-choice     & domain space             & function space \\
modified dual-space  & 40.4     & yes         & ten-choice     & function space           & function space \\
modified dual-space  & 38.2     & yes         & ten-choice     & mono space               & mono space \\
modified dual-space  & 34.8     & yes         & ten-choice     & domain space             & domain space \\
modified dual-space  & 27.5     & yes         & ten-choice     & not used                 & function space \\
modified dual-space  & 25.1     & yes         & ten-choice     & not used                 & mono space \\
modified dual-space  & 14.4     & yes         & ten-choice     & not used                 & domain space \\
\hline
\end{tabular}
} 
\end{center}
\caption {Accuracy with the original five-choice questions and the reformulated ten-choice
questions. In the modified models, we intentionally use the wrong matrix (or no matrix)
for ${\rm sim_d}$ or ${\rm sim_f}$. The modified models show that accuracy decreases when
only one space is used.}
\label{tab:satsum}
\end{table}

\subsubsection{Summary}

The dual-space model performs as well as the current state-of-the-art holistic model and
addresses the issue of linguistic creativity. The results with the reformulated SAT questions
support the claim that there is value in having two separate spaces.

As we mentioned in Section~\ref{subsec:relations}, the task of classifying word pairs
according to their semantic relations \cite{rosario01,rosario02,nastase03} is closely
connected to the problem of measuring relational similarity. \citeA{turney06b} applied
a measure of relational similarity to relation classification by using cosine similarity
as a measure of nearness in a nearest neighbour supervised learning algorithm.
The dual-space model (Equation~\ref{eqn:sat4}) is also suitable for relation classification
with a nearest neighbour algorithm.

\subsection{Similarity of Compositions}
\label{subsec:wordnet-exper}

In this second set of experiments, we apply the dual-space model to noun-modifier
compositions. Given vectors for {\em dog}, {\em house}, and {\em kennel}, we would
like to be able to recognize that {\em dog house} and {\em kennel} are
synonymous. We compare the dual-space model to the holistic approach,
vector addition, and element-wise multiplication. The approaches are evaluated
using multiple-choice questions that are automatically generated from WordNet,
using the WordNet::QueryData Perl interface to WordNet. Table~\ref{tab:kennel} gives
an example of one of the noun-modifier questions.

\begin{table}[h]
\begin{center}
\scalebox{0.92}{
\begin{tabular}{lll}
\hline
Stem:      &       & dog house \\
\hline
Choices:   & (1)   & kennel \\
           & (2)   & dog \\
           & (3)   & house \\
           & (4)   & canine \\
           & (5)   & dwelling \\
           & (6)   & effect \\
           & (7)   & largeness \\
\hline
Solution:  & (1)   & kennel \\
\hline
\end{tabular}
} 
\end{center}
\caption {An example of a multiple-choice noun-modifier composition question.}
\label{tab:kennel}
\end{table}

In these questions, the stem is a bigram and the choices are unigrams.
Choice (1) is the correct answer, (2) is the modifier, and (3)
is the head noun. Choice (4) is a synonym or hypernym of the modifier
and (5) is a synonym or hypernym of the head noun. If no synonyms or hypernyms
can be found, a noun is randomly chosen. The last two choices, (6) and (7),
are randomly selected nouns. Choices (2) and (4) can be either nouns or adjectives,
but the other choices must be nouns.

The stem bigram and the choice unigrams must have corresponding rows in function space
(the space with the least number of rows). The stem bigram must have a noun sense in
WordNet (it may also have senses for other parts of speech). The solution
unigram, (1), must be a member of the synset (synonym set) for the first noun sense of
the stem bigram (the most frequent or dominant sense of the bigram, when the bigram
is used as a noun), and it cannot be simply the hyphenation ({\em dog-house})
or concatenation ({\em doghouse}) of the stem bigram.

These requirements result in a total of 2180 seven-choice questions, which
we randomly split into 680 for training (parameter tuning) and 1500 for
testing.\footnote{The questions are available as an online appendix at
http://jair.org/.} The questions are deliberately designed to be difficult. In particular,
all of the approaches are strongly attracted to choices (2) and (3). Furthermore,
we did not attempt to ensure that the stem bigrams are compositional; some of
them may be idiomatic expressions that no compositional approach could possibly
get right. We did not want to bias the questions by imposing theories
about distinguishing compositions and idioms in their construction.

Let $ab$ represent a noun-modifier bigram ({\em dog house}) and let $c$
represent a unigram ({\em kennel}). We answer the multiple-choice questions
by selecting the unigram that maximizes the compositional similarity,
${\rm sim_c}(ab, c)$, defined as follows:

\begin{align}
\label{eqn:comp1} {\rm sim_1}(ab, c) & =
{\rm geo}({\rm sim_d}(a,c), {\rm sim_d}(b,c), {\rm sim_f}(b,c)) \\
\label{eqn:comp2} {\rm sim_c}(ab, c) & =
\left\{
\begin{array}{rl}
{\rm sim_1}(ab, c) & \mbox{if $a \ne c$ and $b \ne c$} \\
0 & \mbox{otherwise}
\end{array}
\right.
\end{align}

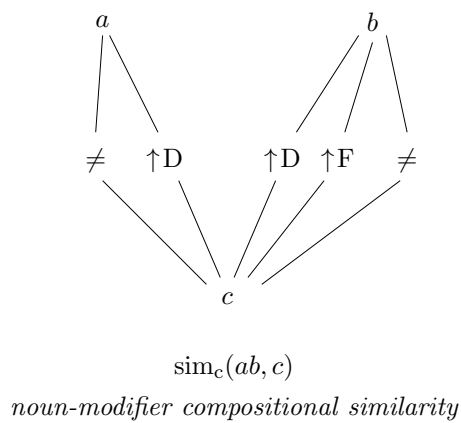
\begin{figure}[h]
\begin{center}
\scalebox{0.92}{
\begin{tikzpicture}
\node at (0.1,5) {$a$};
\node at (4.0,5) {$b$};
\node at (1.9,1) {$c$};
\node at (0.0,3) {$\ne$};
\node at (1.0,3) {$\uparrow \! {\rm D}$};
\node at (2.7,3) {$\uparrow \! {\rm D}$};
\node at (3.5,3) {$\uparrow \! {\rm F}$};
\node at (4.5,3) {$\ne$};
\node at (2.0,0) {${\rm sim_c}(ab, c)$};
\node at (2.0,-0.6) {\em noun-modifier compositional similarity};
\draw (0.2,4.8) -- (0.9,3.4);       
\draw (1.2,2.7) -- (1.8,1.3);       
\draw (3.8,4.8) -- (2.9,3.4);       
\draw (4.0,4.7) -- (3.6,3.4);       
\draw (2.6,2.7) -- (2.0,1.3);       
\draw (3.3,2.7) -- (2.2,1.3);       
\draw (0.1,4.8) -- (0.0,3.4);       
\draw (4.2,4.8) -- (4.5,3.4);       
\draw (0.1,2.7) -- (1.6,1.2);       
\draw (4.4,2.7) -- (2.4,1.2);       
\end{tikzpicture}
} 
\end{center}
\caption{A diagram of Equations \ref{eqn:comp1} and \ref{eqn:comp2}.}
\label{fig:simc}
\end{figure}

\noindent Equations \ref{eqn:comp1} and \ref{eqn:comp2} are illustrated in
Figure~\ref{fig:simc}.

The thinking behind ${\rm sim_1}$ is that $c$ ({\em kennel}$\,$) should have
high domain similarity with both the modifier $a$ ({\em dog}) and the head
noun $b$ ({\em house}); furthermore, the function of the bigram $ab$ ({\em dog house})
is determined by the head noun $b$ ({\em house}), so the head noun should
have high function similarity with $c$ ({\em kennel}$\,$). We add the constraints
$a \ne c$ and $b \ne c$ because ${\rm sim_1}$ by itself tends to have high values
for ${\rm sim_1}(ab, a)$ and ${\rm sim_1}(ab, b)$.\footnote{In spite of these constraints,
it is still worthwhile to include the head noun and the modifier as distractors in the
multiple-choice questions, because it enables us to experimentally evaluate the impact
of these distractors on the various algorithms when the constraints are removed
(see Table~\ref{tab:compconstr}). Also, future users of this dataset
may find a way to avoid these distractors without explicit constraints.} It seems plausible
that humans use constraints like this: We reason that {\em dog house} cannot mean the same
thing as {\em house}, because then the extra word {\em dog} in {\em dog house}
would serve no purpose; it would be meaningless noise.\footnote{In philosophy of language,
\citeA{grice89} argued that proper interpretation of language requires us to
charitably assume that speakers generally do not insert random words into
their speech.}

The constraints $a \ne c$ and $b \ne c$ could be expressed in terms of similarities,
such as ${\rm sim_d}(a,c) < t$ and ${\rm sim_d}(b,c) < t$, where $t$ is a high
threshold (e.g., $t = 0.9$), but this would add another parameter to the model.
We decided to keep the model relatively simple.

When all seven choices for a noun-modifier question have a compositional similarity
of zero, we skip the question. On the training set, the best parameter settings
are $k_{\rm d} = 800$, $p_{\rm d} = 0.3$, $k_{\rm f} = 100$, and $p_{\rm f} = 0.6$.
On the testing set, Equation~\ref{eqn:comp2} correctly answers 874 questions, skips 22
questions, and incorrectly answers 604, yielding an accuracy of 58.3\%.

\subsubsection{Comparison with Other Approaches}

\citeA{mitchell10} compared many different approaches to semantic composition in their
experiments, but they only considered one task (the task we examine in Section~\ref{subsec:phrase-exper}).
In this paper, we have chosen to compare a smaller number of approaches on
a larger number of tasks. We include element-wise multiplication in these
experiments, because this approach had the best performance in \citeS{mitchell10}
experiments. Vector addition is included due to its historical importance and
its simplicity. Although \citeA{mitchell10} found that weighted addition was
better than unweighted addition, we do not include weighted addition in our experiments,
because it did not perform as well as element-wise multiplication in \citeS{mitchell10}
experiments. We include the holistic model as a noncompositional baseline.

Table~\ref{tab:compscores} compares the dual-space model with the holistic
model, element-wise multiplication, and vector addition. For the latter three models,
we try all three spaces.

\begin{table}[h]
\begin{center}
\scalebox{0.92}{
\begin{tabular}{llc}
\hline
Algorithm       & Space                 & Accuracy \\
\hline
dual-space      & domain and function   & 58.3     \\
\hline
holistic        & mono                  & 81.6     \\
holistic        & domain                & 79.1     \\
holistic        & function              & 67.5     \\
\hline
multiplication  & mono                  & 55.7     \\
multiplication  & domain                & 57.5     \\
multiplication  & function              & 46.3     \\
\hline
addition        & mono                  & 48.3     \\
addition        & domain                & 50.1     \\
addition        & function              & 39.8     \\
\hline
\end{tabular}
} 
\end{center}
\caption {Results for the noun-modifier questions.}
\label{tab:compscores}
\end{table}

In this table, {\em dual-space} refers to the dual-space model using Equation~\ref{eqn:comp2}.
In the holistic model, $ab$ is represented by its corresponding row
vector in the given space. Recall from Section~\ref{subsec:building} that, in Step 1.1,
the rows in the matrices correspond to $n$-grams in WordNet, where $n$ may be greater than one.
Thus, for example, {\em dog house} has a corresponding row vector in all three of the spaces.
The holistic model simply uses this row vector as the representation of {\em dog house}. For
element-wise multiplication, $ab$ is represented using Equation~\ref{eqn:mult}. With the
vector addition model, $ab$ is represented by $\mathbf{a} + \mathbf{b}$, where the vectors
are normalized to unit length before they are added. All four models use the constraints
$a \ne c$ and $b \ne c$. All four models use the training data for parameter tuning.

The difference between the dual-space model (58.3\%) and the best variation of element-wise
multiplication (57.5\%) is not statistically significant at the 95\% confidence level,
according to Fisher's Exact Test. However, the difference between the dual-space model
(58.3\%) and the best variation of vector addition (50.1\%) is significant.

\subsubsection{Limitations of the Holistic Approach}

For all three spaces, the holistic model is significantly better than all other models,
but its inability to address the issue of linguistic creativity is a major limitation.
The 2180 multiple-choice questions that we have used in these
experiments were intentionally constructed with the requirement that the stem
bigram must have a corresponding row in function space (see above). This was done so that
we could use the holistic model as a baseline; however, it gives the misleading
impression that the holistic model is a serious competitor with the compositional
approaches. By design, Table~\ref{tab:compscores} shows what the holistic model can achieve
under ideal (but unrealistic) conditions.

\citeS{mitchell10} dataset, used in the experiments in Section~\ref{subsec:phrase-exper},
illustrates the limitations of the holistic model. The dataset consists of 324 distinct
pairs of bigrams, composed of 216 distinct bigrams. Of the 216 bigrams,
28 (13\%) occur in WordNet. Of the 324 pairs of bigrams, 13 (4\%) contain bigrams that
both occur in WordNet. Given the matrices we use here (with rows based on WordNet), the
holistic approach would be reduced to random guessing for 96\% of the pairs
in \citeS{mitchell10} dataset.

It might be argued that the failure of the holistic approach with \citeS{mitchell10} dataset
is due to our decision to base the rows of the matrices on terms from WordNet. However,
suppose we attempt to build a holistic model for all frequent bigrams. The Web 1T \mbox{5-gram}
corpus \cite{brants06} includes a list of all bigrams that appeared 40 or more times in a
terabyte of text, a total of 314,843,401 bigrams. Using a compositional approach, the matrices
we use here can represent the majority of these bigrams. On the other hand, the holistic approach
would require a matrix with 314,843,401 rows, which is considerably beyond the current state of
the art.

One possibility is to build a matrix for the holistic approach as needed, for a given input
set of $n$-grams, instead of building a large, static, multipurpose matrix. There are
two problems with this idea. First, it is slow. \citeA{turney06b} used this approach
for the SAT analogy questions, but it required nine days to run, whereas the dual-space
model can process the SAT questions in a few seconds, given a static, multipurpose matrix.
Second, it requires a large corpus, and the corpus size must grow exponentially
with $n$, the length of the phrases. Longer phrases are more rare, so larger corpora
are needed to gather sufficient data to model the phrases. Larger corpora also result
in longer processing times.

For a given application, it may be wise to have a predefined list of bigrams
with holistic representations, but it would not be wise to expect this list
to be sufficient to cover most bigrams that would be seen in practice.
The creativity of human language use {\em requires} compositional models
\cite{chomsky75,fodor02}. Although the holistic model is included as a
baseline in the experiments, it is not a {\em competitor} for the other models;
it can only {\em supplement} the other models.

\subsubsection{Impact of Constraints}

If we use ${\rm sim_1}$ alone (Equation~\ref{eqn:comp1}), dropping the constraints $a \ne c$ and
$b \ne c$, then accuracy drops signficantly, from 58.3\% to 13.7\%. However, all of the models
benefit greatly from these constraints. In Table~\ref{tab:compconstr}, we take the
best variation of each model from Table~\ref{tab:compscores} and look at what happens
when the constraints are dropped.

\begin{table}[h]
\begin{center}
\scalebox{0.92}{
\begin{tabular}{llccc}
\hline
                &                       & \multicolumn{3}{c}{Accuracy} \\
\cline{3-5}
Algorithm       & Space                 & constraints    & no constraints   & difference \\
\hline
dual-space      & domain and function   & 58.3           & 13.7             & -44.6 \\
holistic        & mono                  & 81.6           & 49.6             & -32.0 \\
multiplication  & domain                & 57.5           & 8.2              & -49.3 \\
addition        & domain                & 50.1           & 2.5              & -47.6 \\
\hline
\end{tabular}
} 
\end{center}
\caption {The impact of the constraints, $a \ne c$ and $b \ne c$, on accuracy.}
\label{tab:compconstr}
\end{table}

\subsubsection{Element-wise Multiplication}

In Section~\ref{subsec:multiplication}, we argued that $\mathbf{c} = \mathbf{a} \odot \mathbf{b}$
is not suitable for row vectors in the matrix $\mathbf{U}_k \mathbf{\Sigma}_k^p$ and we suggested
Equation~\ref{eqn:mult} as an alternative. When we use $\mathbf{c} = \mathbf{a} \odot \mathbf{b}$
with domain space, instead of Equation~\ref{eqn:mult}, performance drops significantly,
from 57.5\% to 21.5\%.

\subsubsection{Impact of Idioms}

Some of the gap between the holistic model and the other models may be due to
idiomatic bigrams in the testing questions. One of the most successful approaches
to determining whether a multiword expression (MWE) is compositional or noncompositional
(idiomatic) is to compare its holistic vector representation with its compositional vector
representation (for example, a high cosine between the two vectors suggests that the
MWE is compositional, not idiomatic) \cite{biemann11,johannsen11}. However, this approach
is not suitable here, because we do not want to {\em assume} that the gap is entirely
due to idiomatic bigrams; instead, we would like to {\em estimate} how much of the
gap is due to idiomatic bigrams.

WordNet contains some clues that we can use as indicators that a bigram might be less
compositional than most bigrams (allowing that compositionality is a matter of degree). One
clue is whether the WordNet gloss of the bigram contains either the head noun or the modifier.
For example, the gloss of {\em dog house} is {\em outbuilding that serves as a shelter for
a dog}, which contains the modifier, {\em dog}. This suggests that {\em dog house} may
be compositional.

We classified each of the 1500 testing set questions as {\em head} (the first five
characters in the {\em head} noun of the bigram match the first five characters
in a word in the bigram's gloss), {\em modifier} (the first five
characters in the {\em modifier} of the bigram match the first five characters
in a word in the bigram's gloss), {\em both} ({\em both} the head and the modifier
match), or {\em neither} ({\em neither} the head nor the modifier match).
The four classes are approximately equally distributed in the testing questions
(424 {\em head}, 302 {\em modifier}, 330 {\em both}, and 444 {\em neither}).
We match on the first five characters to allow for cases like {\em brain surgeon},
which has the gloss {\em someone who does surgery on the nervous system (especially
the brain)}. This bigram is classified as {\em both}, because the first five
characters of {\em surgeon} match the first five characters of {\em surgery}.

Table~\ref{tab:gloss} shows how the accuracy of the models varies over the four classes
of questions. For the three compositional models (dual-space, multiplication, addition),
the {\em neither} class is significantly less accurate than the other three classes
(Fisher's Exact Test, 95\% confidence), but the difference is not significant for the
holistic model. For the three compositional models, the {\em neither} class is 17\% to 20\%
less accurate than the other classes. This supports the view that a significant fraction of
the wrong answers of the compositional models are due to noncompositional bigrams.

\begin{table}[h]
\begin{center}
\scalebox{0.92}{
\begin{tabular}{llccccc}
\hline
                &                       & \multicolumn{5}{c}{Accuracy} \\
\cline{3-7}
Algorithm       & Space                 & both    & head   & modifier & neither & all  \\
\hline
dual-space      & domain and function   & 63.0    & 63.0   & 64.6     & 45.9    & 58.3 \\
holistic        & mono                  & 82.7    & 83.7   & 82.1     & 78.4    & 81.6 \\
multiplication  & domain                & 61.8    & 63.7   & 62.9     & 44.8    & 57.5 \\
addition        & domain                & 53.6    & 56.8   & 56.3     & 36.7    & 50.1 \\
\hline
\end{tabular}
} 
\end{center}
\caption {The variation of accuracy for different classes of bigram glosses.}
\label{tab:gloss}
\end{table}

Another clue for compositionality in WordNet is whether the head noun is a hypernym
of the bigram. For example, {\em surgeon} is a hypernym of {\em brain surgeon}.
We classified each of the 1500 testing set questions as {\em hyper} (the head noun
is a member of the synset of the immediate hypernym for the first noun sense of the
bigram; we do not look further up in the hypernym hierarchy and we do not look at
other senses of the bigram) or {\em not} (not {\em hyper}). In the testing set,
621 questions are {\em hyper} and 879 are {\em not}.

Table~\ref{tab:hyper} gives the accuracy of the models for each of the classes.
This table has the same general pattern as Table~\ref{tab:gloss}. The three
compositional models have significantly lower accuracy for the {\em not} class,
with decreases from 6\% to 8\%. There is no significant difference for the holistic
model.

\begin{table}[h]
\begin{center}
\scalebox{0.92}{
\begin{tabular}{llccc}
\hline
                &                       & \multicolumn{3}{c}{Accuracy} \\
\cline{3-5}
Algorithm       & Space                 & hyper   & not    & all  \\
\hline
dual-space      & domain and function   & 62.0    & 55.6   & 58.3 \\
holistic        & mono                  & 81.0    & 82.0   & 81.6 \\
multiplication  & domain                & 61.8    & 54.5   & 57.5 \\
addition        & domain                & 54.8    & 46.8   & 50.1 \\
\hline
\end{tabular}
} 
\end{center}
\caption {The variation of accuracy for different classes of bigram hypernyms.}
\label{tab:hyper}
\end{table}

\subsubsection{Order Sensitivity}

Note that vector addition and element-wise multiplication lack order sensitivity, but
Equation~\ref{eqn:comp2} is sensitive to order, ${\rm sim_c}(ab, c) \ne {\rm sim_c}(ba, c)$.
We can see the impact of this by reformulating the noun-modifier questions so that they
test for order-sensitivity. First we expand each choice unigram $c$ by including the
stem bigram $ab$, resulting in the explicit comparison $ab \sim c$. For each expanded
choice, $ab \sim c$, we then generate another choice, $ba \sim c$. This increases the
number of choices from seven to fourteen. Due to symmetry, vector addition and element-wise
multiplication must assign the same similarity to both $ab \sim c$ and $ba \sim c$.

Table~\ref{tab:reform} compares the dual-space model with 
element-wise multiplication and vector addition, using the reformulated fourteen-choice
noun-modifier questions. The holistic model is not included in this table because
there are no rows in the matrices for the reversed $ba$ bigrams (which may be seen
as another illustration of the limits of the holistic model). On this stricter
test, the dual-space model is significantly more accurate than both element-wise
multiplication and vector addition (Fisher's Exact Test, 95\% confidence).

\begin{table}[h]
\begin{center}
\scalebox{0.92}{
\begin{tabular}{llc}
\hline
Algorithm           & Space                 & Accuracy \\
\hline
dual-space          & domain and function   & 41.5 \\
multiplication      & domain                & 27.4 \\
modified dual-space & function alone        & 25.7 \\
modified dual-space & domain alone          & 25.7 \\
addition            & domain                & 22.5 \\
\hline
\end{tabular}
} 
\end{center}
\caption {Results for the reformulated fourteen-choice noun-modifier questions.}
\label{tab:reform}
\end{table}

For the dual-space model to perform well with the fourteen-choice questions, we need
both ${\rm sim_d}$ and ${\rm sim_f}$. If we drop ${\rm sim_d}$ from Equation~\ref{eqn:comp2}
({\em function alone} in Table~\ref{tab:reform}), then we are ignoring the modifier and
only paying attention to the head noun. Accuracy drops from 41.5\% down to 25.7\%. If we
drop ${\rm sim_f}$ from Equation~\ref{eqn:comp2} ({\em domain alone} in Table~\ref{tab:reform}),
then the equation becomes symmetrical, so the same similarity is assigned to both
$ab \sim c$ and $ba \sim c$. Accuracy drops from 41.5\% down to 25.7\%.\footnote{It
is only a coincidence that both modified dual-space models have an accuracy of 25.7\%
on the fourteen-choice questions. Although their aggregate accuracy is the same, on individual
questions, the two models typically select different choices.} The dual-space model is
significantly more accurate than either of these modified dual-space models (Fisher's Exact
Test, 95\% confidence).

\subsubsection{Summary}

With the reformulated fourteen-choice noun-modifier questions (Table~\ref{tab:reform}),
the dual-space is significantly better than element-wise multiplication and vector addition.
With the original seven-choice questions (Table~\ref{tab:compscores}), the difference is not
as large, because these questions do not test for order. Unlike element-wise multiplication
and vector addition, the dual-space model addresses the issue of order sensitivity. Unlike
the holistic model, the dual-space addresses the issue of linguistic creativity.

\subsection{Similarity of Phrases}
\label{subsec:phrase-exper}

In this subsection, we apply the dual-space model to measuring the similarity
of phrases, using \citeS{mitchell10} dataset of human similarity ratings
for pairs of phrases. The dataset includes three types of phrases,
adjective-noun, noun-noun, and verb-object. There are 108 pairs of each type
($108 \times 3 = 324$ pairs of phrases). Each pair of phrases was rated by 18 human
subjects. The ratings use a 7 point scale, in which 1 signifies the lowest degree of
similarity and 7 signifies the highest degree. Table~\ref{tab:mitlap1} gives some examples.

\begin{table}[h]
\begin{center}
\scalebox{0.92}{
\begin{tabular}{clclc}
\hline
Participant  & Phrase type & Group & Phrase pair & Similarity \\
\hline
114 & adjective-noun & 2 & certain circumstance $\sim$ particular case   & 6 \\
114 & adjective-noun & 2 & large number $\sim$ great majority            & 4 \\
114 & adjective-noun & 2 & further evidence $\sim$ low cost              & 2 \\
\hline
109 & noun-noun      & 0 & environment secretary $\sim$ defence minister & 6 \\
109 & noun-noun      & 0 & action programme $\sim$ development plan      & 4 \\
109 & noun-noun      & 0 & city centre $\sim$ research work              & 1 \\
\hline
111 & verb-object    & 2 & lift hand $\sim$ raise head                   & 7 \\
111 & verb-object    & 2 & satisfy demand $\sim$ emphasise need          & 4 \\
111 & verb-object    & 2 & like people $\sim$ increase number            & 1 \\
\hline
\end{tabular}
} 
\end{center}
\caption {Examples of phrase pair similarity ratings from \citeS{mitchell10} dataset.
Similarity ratings vary from 1 (lowest) to 7 (highest).}
\label{tab:mitlap1}
\end{table}

Let $ab$ represent the first phrase in a pair of phrases ({\em environment secretary})
and let $cd$ represent the second phrase ({\em defence minister}). We rate the
similarity of the phrase pairs by ${\rm sim_p}(ab, cd)$, defined as follows:

\begin{equation}
\label{eqn:simp}
{\rm sim_p}(ab, cd)
= {\rm geo}({\rm sim_d}(a,c), {\rm sim_d}(b,d), {\rm sim_f}(a,c), {\rm sim_f}(b,d))
\end{equation}

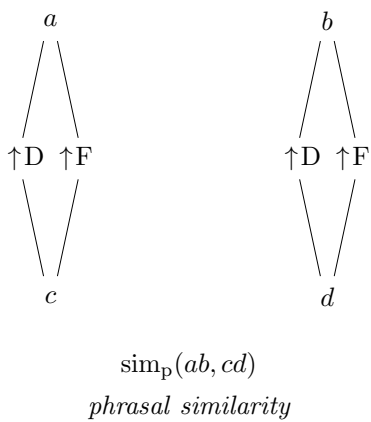
\begin{figure}[h]
\begin{center}
\scalebox{0.92}{
\begin{tikzpicture}
\node at (0.4,5) {$a$};
\node at (4.4,5) {$b$};
\node at (0.4,1) {$c$};
\node at (4.4,1) {$d$};
\node at (0.4,3) {$\uparrow \! {\rm D} \; \uparrow \! {\rm F}$};
\node at (4.4,3) {$\uparrow \! {\rm D} \; \uparrow \! {\rm F}$};
\node at (2.4,0) {${\rm sim_p}(ab, cd)$};
\node at (2.4,-0.6) {\em phrasal similarity};
\draw (0.0,3.3) -- (0.3,4.7);     
\draw (0.8,3.3) -- (0.5,4.7);     
\draw (4.0,3.3) -- (4.3,4.7);     
\draw (4.8,3.3) -- (4.5,4.7);     
\draw (0.3,1.3) -- (0.0,2.7);     
\draw (0.5,1.3) -- (0.8,2.7);     
\draw (4.3,1.3) -- (4.0,2.7);     
\draw (4.5,1.3) -- (4.8,2.7);     
\end{tikzpicture}
} 
\end{center}
\caption{A diagram of Equation \ref{eqn:simp}.}
\label{fig:simp}
\end{figure}

\noindent This equation is based on the instructions to the
human participants \cite[Appendix B]{mitchell10}, which imply that both
function and domain similarity must be high for a phrase pair to get a high
similarity rating. Figure~\ref{fig:simp} illustrates the reasoning behind this
equation. We want high domain and function similarities between the corresponding
components of the phrases $ab$ and $cd$.

\subsubsection{Experimental Setup}

\citeA{mitchell10} divided their dataset into a development set (for tuning
parameters) and an evaluation set (for testing the tuned models). The development
set has 6 ratings for each phrase pair and the evaluation set has 12 ratings for
each phrase pair. The development and evaluation sets contain the same phrase pairs, but
with judgments by different participants. Thus there are $6 \times 324 = 1,944$
rated phrase pairs in the development set and $12 \times 324 = 3,888$ ratings in the
evaluation set.\footnote{The information in this paragraph is based on Section~4.3
of the paper by \citeA{mitchell10} and personal communication with Jeff Mitchell in June,
2010.}

For a more challenging evaluation, we divide the dataset by phrase pairs rather than
by participants. Our development set has 108 phrase pairs with 18 ratings each
and the evaluation set has 216 phrase pairs with 18 ratings each. For each of the three
phrase types, we randomly select 36 phrase pairs for the development set
($3 \times 36 = 108$ phrase pairs) and 72 for the evaluation set ($3 \times 72 = 216$
phrase pairs). Thus there are $18 \times 108 = 1,944$ ratings in the development set
and $18 \times 216 = 3,888$ in the evaluation set.

\citeA{mitchell10} use Spearman's rank correlation coefficient (Spearman's rho)
to evaluate the performance of various vector composition algorithms
on the task of emulating the human similarity ratings. For a given phrase
type, the 108 phrase pairs are divided into 3 groups of 36 pairs each.
For each group in the evaluation set, 12 people gave similarity ratings to the
pairs in the given group. Each group of 36 pairs was given to a different
group of 12 people. The score of an algorithm for a given phrase type
is the average of three rho values, one rho for each of the three groups.
With 12 people rating 36 pairs in a group, there are $12 \times 36 = 432$
ratings. These human ratings are represented by a vector of 432 numbers.
An algorithm only generates one rating for each pair in a group, yielding
36 numbers. To make the algorithm's ratings comparable to the human ratings,
the algorithm's ratings are duplicated 12 times, yielding a vector of 432
numbers. Spearman's rho is then calculated with these two vectors of 432
ratings. For 3 phrase types with 3 rho values each and 432 ratings per rho value, we
have 3,888 ratings.\footnote{The information in this paragraph is based on personal
communication with Jeff Mitchell in June, 2010. \citeS{mitchell10} paper
does not describe how Spearman's rho is applied.}

We believe that this evaluation method underestimates the performance of
the algorithms. Combining ratings from different people into one vector
of 432 numbers does not allow the correlation to adapt to different
biases. If one person gives consistently low ratings and another person
gives consistently high ratings, but both people have the same ranking,
and this ranking matches the algorithm's ranking, then the algorithm
should get a high score. For a more fair evaluation, we
score an algorithm by calculating one rho value for each human participant
for the given phrase type, and then we calculate the average of the
rho values for all of the participants.

For a given phrase type, the 108 phrase pairs are divided into 3 groups of
36 pairs each. For the development set, we randomly select 12 phrase pairs
from each of the 3 groups ($3 \times 12 = 36$ phrase pairs per phrase type).
This leaves 24 phrase pairs in each of the 3 groups for the evaluation set
($3 \times 24 = 72$ phrase pairs per phrase type). Each human participant's
ratings are represented by a vector of 24 numbers. An algorithm's
ratings are also represented by a vector of 24 numbers. A rho value is
calculated with these two vectors of 24 numbers as input. For a given
phrase type, the algorithm's score is the average of 54 rho values (18 participants
per group $\times$ 3 groups = 54 rho values). For 3 phrase types with 54 rho values
each and 24 ratings per rho value, we have 3,888 ratings.

\subsubsection{Comparison with Other Approaches}

Table~\ref{tab:method1} compares the dual-space model to vector addition and
element-wise multiplication. We use the development set to tune the parameters
for all three approaches. For vector addition, $ab$ is represented by
$\mathbf{a} + \mathbf{b}$ and $cd$ is represented by $\mathbf{c} + \mathbf{d}$.
The similarity of $ab$ and $cd$ is given by the cosine of the
two vectors. Element-wise multiplication uses Equation~\ref{eqn:mult}
to represent $ab$ and $cd$. The dual-space model uses Equation~\ref{eqn:simp}.

\begin{table}[h]
\begin{center}
\scalebox{0.92}{
\begin{tabular}{llllll}
\hline
               & \multicolumn{4}{c}{Correlation for each phrase type} & \\
\cline{2-5}
Algorithm      & ad-nn  & nn-nn  & vb-ob  & avg   & Comment \\
\hline
human          & 0.56   & 0.54   & 0.57   & 0.56  & leave-one-out correlation between subjects \\
\hline
dual-space     & 0.48   & 0.54   & 0.43   & 0.48  & domain and function space \\
\hline
addition       & 0.47   & 0.61   & 0.42   & 0.50  & mono space \\
addition       & 0.32   & 0.55   & 0.41   & 0.42  & domain space \\
addition       & 0.49   & 0.55   & 0.48   & 0.51  & function space \\
\hline
multiplication & 0.43   & 0.57   & 0.41   & 0.47  & mono space \\
multiplication & 0.35   & 0.58   & 0.39   & 0.44  & domain space \\
multiplication & 0.39   & 0.45   & 0.27   & 0.37  & function space \\
\hline
\end{tabular}
} 
\end{center}
\caption {Performance of the models on the evaluation dataset.}
\label{tab:method1}
\end{table}

The average correlation of the dual-space model (0.48) is significantly
below the average correlation of vector addition using function space (0.51).
Element-wise multiplication with mono space (0.47) is also significantly
below vector addition using function space (0.51). The difference between
the dual-space model (0.48) and element-wise multiplication with mono space (0.47)
is not signficant. The average correlation for an algorithm is based on 162 rho
values (3 phrase types $\times$ 3 groups $\times$ 18 participants = 162 rho values
= 162 participants). We calculate the statistical significance using a paired t-test
with a 95\% significance level, based on 162 pairs of rho values.

\subsubsection{Order Sensitivity}

\citeS{mitchell10} dataset does not test for order sensitivity.
Given a phrase pair $ab \sim cd$, we can test for order sensitivity
by adding a new pair $ab \sim dc$. We assume that all such new pairs
would be given a rating of 1 by the human participants. In Table~\ref{tab:mitlap2},
we show what happens when this transformation is applied to the
examples in Table~\ref{tab:mitlap1}. To save space, we only give the
examples for participant number 114.

\begin{table}[h]
\begin{center}
\scalebox{0.92}{
\begin{tabular}{clclc}
\hline
Participant  & Phrase type & Group & Phrase pair & Similarity \\
\hline
114 & adjective-noun & 2 & certain circumstance $\sim$ particular case   & 6 \\
114 & adjective-noun & 2 & certain circumstance $\sim$ case particular   & 1 \\
114 & adjective-noun & 2 & large number $\sim$ great majority            & 4 \\
114 & adjective-noun & 2 & large number $\sim$ majority great            & 1 \\
114 & adjective-noun & 2 & further evidence $\sim$ low cost              & 2 \\
114 & adjective-noun & 2 & further evidence $\sim$ cost low              & 1 \\
\hline
\end{tabular}
} 
\end{center}
\caption {Testing for order sensitivity by adding new phrase pairs.}
\label{tab:mitlap2}
\end{table}

Table~\ref{tab:method2} gives the results with the new, expanded
dataset. With this more stringent dataset, the dual-space model performs significantly
better than both vector addition and vector multiplication.
Unlike element-wise multiplication and vector addition, the dual-space model
addresses the issue of order sensitivity.

\begin{table}[h]
\begin{center}
\scalebox{0.92}{
\begin{tabular}{llllll}
\hline
               & \multicolumn{4}{c}{Correlation for each phrase type} & \\
\cline{2-5}
Algorithm      & ad-nn  & nn-nn  & vb-ob  & avg   & Comment \\
\hline
human          & 0.71   & 0.81   & 0.73   & 0.75  & leave-one-out correlation between subjects \\
\hline
dual-space     & 0.66   & 0.37   & 0.62   & 0.55  & domain and function space \\
\hline
addition       & 0.22   & 0.25   & 0.19   & 0.22  & mono space \\
addition       & 0.15   & 0.22   & 0.18   & 0.18  & domain space \\
addition       & 0.23   & 0.23   & 0.19   & 0.22  & function space \\
\hline
multiplication & 0.20   & 0.24   & 0.18   & 0.21  & mono space \\
multiplication & 0.18   & 0.22   & 0.18   & 0.19  & domain space \\
multiplication & 0.18   & 0.19   & 0.12   & 0.17  & function space \\
\hline
\end{tabular}
} 
\end{center}
\caption {Performance when the dataset is expanded to test for order sensitivity.}
\label{tab:method2}
\end{table}

We manually inspected the new pairs that were automatically rated 1 and found
that a rating of 1 was reasonable in all cases, although some cases could be
disputed. For example, the original noun-noun pair {\em tax charge $\sim$ interest rate}
generates the new pair {\em tax charge $\sim$ rate interest} and the original
verb-object pair {\em produce effect $\sim$ achieve result} generates the new
pair {\em produce effect $\sim$ result achieve}. It seems that we have a
natural tendency to correct these incorrectly ordered pairs in our minds and
then assign them higher ratings than they deserve. We predict that human
ratings of these new pairs would vary greatly, depending on the instructions
that were given to the human raters. If the instructions emphasized the
importance of word order, the new pairs would get low ratings. This prediction
is supported by the results of SemEval 2012 Task 2 \cite{jurgens12}, where the instructions
to the raters emphasized the importance of word order and wrongly ordered pairs received
low ratings.

\subsubsection{Summary}

When the dataset does not test for order sensitivity, vector addition
performs slightly better than the dual-space model. When the dataset tests
for order sensitivity, the dual-space model surpasses both vector addition and
element-wise multiplication by a large margin.

\subsection{Domain versus Function as Associated versus Similar}
\label{subsec:chiarello-exper}

\citeA{chiarello90} created a dataset of 144 word pairs that they labeled {\em similar-only},
{\em associated-only}, or {\em similar+associated} (48 pairs in each of the three classes).
Table~\ref{tab:associated} shows some examples from their dataset. These labeled pairs were
created for cognitive psychology experiments with human subjects. In their experiments, they
found evidence  that processing {\em associated} words engages the left and right hemispheres
of the brain in ways that are different from processing {\em similar} words. That is, it seems
that there is a fundamental neurological difference between these two types of semantic
relatedness.\footnote{There is some controversy among cognitive scientists over the
distinction between semantic similarity and association \cite{mcrae11}.}

\begin{table}[h]
\begin{center}
\scalebox{0.92}{
\begin{tabular}{ll}
\hline
Word pair & Class label \\
\hline
table:bed      & similar-only \\
music:art      & similar-only \\
hair:fur       & similar-only \\
house:cabin    & similar-only \\
\hline
cradle:baby    & associated-only \\
mug:beer       & associated-only \\
camel:hump     & associated-only \\
cheese:mouse   & associated-only \\
\hline
ale:beer       & similar+associated \\
uncle:aunt     & similar+associated \\
pepper:salt    & similar+associated \\
frown:smile    & similar+associated \\
\hline
\end{tabular}
} 
\end{center}
\caption {Examples of word pairs from \citeA{chiarello90}, labeled {\em similar-only},
{\em associated-only}, or {\em similar+associated}. The full dataset is in their Appendix.}
\label{tab:associated}
\end{table}

We hypothesize that similarity in domain space, ${\rm sim_d}(a,b)$, is a measure of the
degree to which two words are {\em associated} and similarity in function space,
${\rm sim_f}(a,b)$, is a measure of the degree to which two words are {\em similar}.
To test this hypothesis, we define similar-only, ${\rm sim_{so}}(a,b)$,
associated-only, ${\rm sim_{ao}}(a,b)$, and similar+associated, ${\rm sim_{sa}}(a,b)$,
as follows:

{\allowdisplaybreaks 
\begin{align}
{\rm ratio}(x, y) & =
\left\{
\begin{array}{rl}
x / y & \mbox{if $x > 0$ and $y > 0$} \\
0 & \mbox{otherwise}
\end{array}
\right. \\
\label{eqn:simso} {\rm sim_{so}}(a, b) & = {\rm ratio}({\rm sim_f}(a, b) , {\rm sim_d}(a, b)) \\
\label{eqn:simao} {\rm sim_{ao}}(a, b) & = {\rm ratio}({\rm sim_d}(a, b) , {\rm sim_f}(a, b)) \\
\label{eqn:simsa} {\rm sim_{sa}}(a, b) & = {\rm geo}({\rm sim_d}(a, b) , {\rm sim_f}(a, b))
\end{align}
} 

\begin{figure}[h]
\begin{center}
\scalebox{0.92}{
\begin{tikzpicture}
\node at (1,5)  {$a$};
\node at (5,5)  {$a$};
\node at (9,5)  {$a$};
\node at (1,1)  {$b$};
\node at (5,1)  {$b$};
\node at (9,1)  {$b$};
\node at (1,3) {$\downarrow \! {\rm D} \; \uparrow \! {\rm F}$};
\node at (5,3) {$\uparrow \! {\rm D} \; \downarrow \! {\rm F}$};
\node at (9,3) {$\uparrow \! {\rm D} \; \uparrow \! {\rm F}$};
\node at (1,0) {${\rm sim_{so}}(a, b)$};
\node at (5,0) {${\rm sim_{ao}}(a, b)$};
\node at (9,0) {${\rm sim_{sa}}(a, b)$};
\node at (1,-0.6) {\em similar-only};
\node at (5,-0.6) {\em associated-only};
\node at (9,-0.6) {\em similar+associated};
\draw (0.9,4.7) -- (0.6,3.3);     
\draw (1.1,4.7) -- (1.4,3.3);     
\draw (0.6,2.7) -- (0.9,1.3);     
\draw (1.4,2.7) -- (1.1,1.3);     
\draw (4.9,4.7) -- (4.6,3.3);     
\draw (5.1,4.7) -- (5.4,3.3);     
\draw (4.6,2.7) -- (4.9,1.3);     
\draw (5.4,2.7) -- (5.1,1.3);     
\draw (8.9,4.7) -- (8.6,3.3);     
\draw (9.1,4.7) -- (9.4,3.3);     
\draw (8.6,2.7) -- (8.9,1.3);     
\draw (9.4,2.7) -- (9.1,1.3);     
\end{tikzpicture}
} 
\end{center}
\caption{Diagrams of Equations \ref{eqn:simso}, \ref{eqn:simao}, and \ref{eqn:simsa}.}
\label{fig:3sims}
\end{figure}
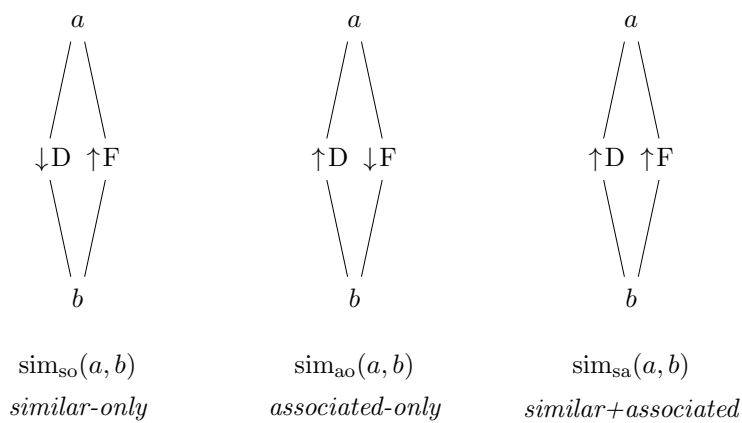

\noindent The intention is that ${\rm sim_{so}}$ is high when ${\rm sim_f}$ is high
and ${\rm sim_d}$ is low, ${\rm sim_{ao}}$ is high when ${\rm sim_d}$ is high
and ${\rm sim_f}$ is low, and ${\rm sim_{sa}}$ is high when both ${\rm sim_d}$
and ${\rm sim_f}$ are high. This is illustrated in Figure~\ref{fig:3sims}.

\subsubsection{Evaluation}

From the experiments in the three preceding subsections, we have three sets
of parameter settings for the dual-space model. Table~\ref{tab:params} shows
these parameter values. In effect, these three sets of parameter setttings
give us three variations of the similarity measures, ${\rm sim_{so}}$,
${\rm sim_{ao}}$, and ${\rm sim_{sa}}$. We will evaluate the three variations
to see how well they correspond to the labels in \citeS{chiarello90} dataset.

\begin{table}[h]
\begin{center}
\scalebox{0.92}{
\begin{tabular}{llccccc}
\hline
Similarity & Description & Section & $k_{\rm d}$ & $p_{\rm d}$ & $k_{\rm f}$ & $p_{\rm f}$ \\
\hline
${\rm sim_r}(a\!:\!b, c\!:\!d)$ & similarity of relations &
\ref{subsec:sat-exper} & 800 & -0.1 & 300 & 0.5 \\
${\rm sim_c}(ab, c)$ & similarity of noun-modifier compositions &
\ref{subsec:wordnet-exper} & 800 &  0.3 & 100 & 0.6 \\
${\rm sim_p}(ab, cd)$ & similarity of phrases &
\ref{subsec:phrase-exper}  & 200 &  0.3 & 600 & 0.6 \\
\hline
\end{tabular}
} 
\end{center}
\caption {Parameter settings for the dual-space model.}
\label{tab:params}
\end{table}

For a given similarity measure, such as ${\rm sim_{so}}$, we can sort the
144 word pairs in descending order of their similarities and then look at
the top $N$ pairs to see how many of them have the desired label; in the
case of ${\rm sim_{so}}$, we would like to see that the majority of the
top $N$ have the label {\em similar-only}. Table~\ref{tab:topn} shows the
percentage of pairs that have the desired labels for each of the three
variations of the three similarity measures. Note that random guessing
would yield 33\%, since the three classes of pairs have the same size.

\begin{table}[h]
\begin{center}
\scalebox{0.92}{
\begin{tabular}{lcccc}
\hline
                     &       & \multicolumn{3}{c}{Percentage of top $N$ with desired label} \\
\cline{3-5}
Source of parameters & $N$   & similar-only          & associated-only       & similar+associated \\
\hline
${\rm sim_r}(a\!:\!b, c\!:\!d)$ & 10  & 70  & 90  & 90 \\
                                & 20  & 80  & 85  & 80 \\
                                & 30  & 63  & 77  & 73 \\
\hline
${\rm sim_c}(ab, c)$            & 10  & 90  & 90  & 80 \\
                                & 20  & 80  & 70  & 70 \\
                                & 30  & 70  & 67  & 73 \\
\hline
${\rm sim_p}(ab, cd)$           & 10  & 50  & 90  & 80 \\
                                & 20  & 65  & 80  & 80 \\
                                & 30  & 47  & 77  & 73 \\
\hline
\end{tabular}
} 
\end{center}
\caption {Percentage of the top $N$ word pairs with the desired labels.}
\label{tab:topn}
\end{table}

For all three sets of parameter settings, Table~\ref{tab:topn} displays a
high density of the desired labels at the tops of the sorted lists. The density
slowly decreases as we move down the lists. This is evidence that the three
similarity measures are capturing the three classes of \citeA{chiarello90}.

As another test of the hypothesis, we use the three similarity measures to
create feature vectors of three elements for each word pair. That is,
the word pair $a\!:\!b$ is represented by the feature vector
$\langle {\rm sim_{so}}(a,b), {\rm sim_{ao}}(a,b), {\rm sim_{sa}}(a,b) \rangle$.
We then use supervised learning with ten-fold cross-validation to classify
the feature vectors into the three classes of \citeA{chiarello90}. For the
learning algorithm, we use logistic regression, as implemented in
Weka.\footnote{Weka is available at http://www.cs.waikato.ac.nz/ml/weka/.}
The results are summarized in Table~\ref{tab:weka}. These results lend
further support to the hypothesis that similarity in domain space, ${\rm sim_d}(a,b)$,
is a measure of the degree to which two words are {\em associated} and similarity in
function space, ${\rm sim_f}(a,b)$, is a measure of the degree to which two words are
{\em similar}.

\begin{table}[h]
\begin{center}
\scalebox{0.92}{
\begin{tabular}{lccccc}
\hline
                     &          & \multicolumn{4}{c}{F-measure} \\
\cline{3-6}
Source of parameters & Accuracy & similar-only & associated-only & similar+associated & average \\
\hline
${\rm sim_r}(a\!:\!b, c\!:\!d)$ & 61.1 & 0.547 & 0.660 & 0.625 & 0.611 \\
${\rm sim_c}(ab, c)$            & 59.0 & 0.583 & 0.702 & 0.490 & 0.592 \\
${\rm sim_p}(ab, cd)$           & 58.3 & 0.472 & 0.699 & 0.563 & 0.578 \\
\hline
\end{tabular}
} 
\end{center}
\caption {Performance of logistic regression with the three similarity measures
as features.}
\label{tab:weka}
\end{table}

In Table~\ref{tab:topn}, similar-only seems more sensitive to the parameter settings
than associated-only and similar+associated. We hypothesize that this is because function
similarity is more difficult to measure than domain similarity. Note that the construction
of function space (Section~\ref{subsec:function}) is more complex than the construction of
domain space (Section~\ref{subsec:domain}). Intuitively, it seems easier to identify
the domain of a thing than to identify its functional role. \citeS{gentner91} work
suggests that children master domain similarity before they become competent with
function similarity.

\section{Discussion of Experiments}
\label{sec:discussion}

This section discusses the results of the previous section.

\subsection{Summary of Results}
\label{subsec:expersum}

In Section~\ref{subsec:sat-exper}, we used 374 multiple-choice analogy questions
to evaluate the dual-space model of relational similarity, ${\rm sim_r}(a\!:\!b, c\!:\!d)$.
The difference between the performance of the dual-space model (51.1\% accuracy)
and the best past result (56.1\% accuracy), using a holistic model, was not
statistically significant. Experiments with a reformulated version of the questions,
designed to test order sensitivity, supported the hypothesis that both domain and
function space are required. Function space by itself is not sensitive to order
and merging the two spaces (mono space) causes a significant drop in performance.

In Section~\ref{subsec:wordnet-exper}, we automatically generated 2,180 multiple-choice
noun-modifier composition questions with WordNet, to evaluate the dual-space model of
noun-modifier compositional similarity, ${\rm sim_c}(ab, c)$. The difference between the
performance of the dual-space model (58.3\% accuracy) and the state-of-the-art
element-wise multiplication model (57.5\% accuracy) was not statistically significant.
The best performance was obtained with a holistic model (81.6\%), but this
model does not address the issue of linguistic creativity. Further experiments
suggest that a significant fraction of the gap between the holistic model and
the other models is due to noncompositional phrases. A limitation of the element-wise
multiplication model is lack of sensitivity to order. Experiments with a reformulated version
of the questions, designed to test order sensitivitiy, demonstrated a statistically
significant advantage to the dual-space model over the element-wise multiplication
and vector addition models.

In Section~\ref{subsec:phrase-exper}, we used \citeS{mitchell10} dataset of 324
pairs of phrases to evaluate the dual-space model of phrasal similarity,
${\rm sim_p}(ab, cd)$. A reformulated version of the dataset, modified to test order
sensitivitiy, showed a statistically significant advantage to the dual-space model over
the element-wise multiplication and vector addition models.

In Section~\ref{subsec:chiarello-exper}, we used \citeS{chiarello90} dataset of
144 word pairs, labeled {\em similar-only}, {\em associated-only}, or
{\em similar+associated}, to test the hypothesis that similarity in domain space,
${\rm sim_d}(a,b)$, is a measure of the degree to which two words are {\em associated}
and similarity in function space, ${\rm sim_f}(a,b)$, is a measure of the degree to
which two words are {\em similar}. The experimental results support the hypothesis.
This is interesting because \citeA{chiarello90} argue that there is a
fundamental neurological difference in the way people process these two kinds
of semantic relatedness.

The experiments support the claim that the dual-space model can address the
issues of linguistic creativity, order sensitivity, and adaptive capacity.
Furthermore, the dual-space model provides a unified
approach to both semantic relations and semantic composition.

\subsection{Corpus-based Similarity versus Lexicon-based Similarity}
\label{subsec:lexicon}

The results in Section~\ref{subsec:chiarello-exper} suggest that function similarity
may correspond to the kind of taxonomical similarity that is often associated with
lexicons, such as WordNet \cite{resnik95,jiang97,leacock98,hirst98}. The word
pairs in Table~\ref{tab:associated} that are labeled {\em similar-only} are
the kinds of words that typically share a common hypernym in a taxonomy. For
example, {\em table:bed} share the hypernym {\em furniture}. We believe that
this is correct, but it does not necessarily imply that lexicon-based similarity
measures would be better than a corpus-based approach, such as we have used here.

Of the various similarities in Section~\ref{sec:experiments}, arguably
relational similarity, \mbox{${\rm sim_r}(a\!:\!b, c\!:\!d)$}, makes the most use
of function similarity. By itself, function similarity achieves 50.8\% on
the SAT questions (original five-choice version; see Table~\ref{tab:satsum}).
However, the best performance achieved on the SAT questions using WordNet
is 43.0\% \cite{veale04}. The difference is statistically significant
at the 95\% confidence level, based on Fisher's Exact Test.

Consider the analogy {\em traffic} is to {\em street} as {\em water} is to
{\em riverbed}. One of the SAT questions involves this analogy, with
{\em traffic}$\,:${\em street} as the stem pair and {\em water}$\,:${\em riverbed}
as the correct choice. Both ${\rm sim_r}(a\!:\!b, c\!:\!d)$ (Equation~\ref{eqn:sat4})
and function similarity by itself (Equation~\ref{eqn:sat1}) make the correct
choice. We can recognize that {\em traffic} and {\em water} have a high
degree of function similarity; in fact, this similarity is used in hydrodynamic
models of traffic flow \cite{daganzo94}. However, we must climb the WordNet hierachy all the
way up to {\em entity} before we find a shared hypernym for {\em traffic} and {\em water}.
We believe that no manually generated lexicon can capture all of the functional
similarity that can be discovered in a large corpus.

\section{Theoretical Considerations}
\label{sec:theory}

This section examines some theoretical questions about the dual-space model.

\subsection{Vector Composition versus Similarity Composition}
\label{subsec:vec-vs-sim}

In the dual-space model, a phrase has no stand-alone, general-purpose representation, as a composite
phrase, apart from the representations of the component words. The composite meaning is
constructed in the context of a given task. For example, if the task is to measure the
similarity of the relation in {\em dog}$\,:${\em house} to the relation in
{\em bird}$\,:${\em nest}, then we compose the meanings of {\em dog} and {\em house} one
way (see Section~\ref{subsec:sat-exper}); if the task is to measure the similarity
of the phrase {\em dog house} to the word {\em kennel}, then we compose the meanings of
{\em dog} and {\em house} another way (see Section~\ref{subsec:wordnet-exper}); if the task is
to measure the similarity of the phrase {\em dog house} to the phrase {\em canine shelter},
then we compose the meanings of {\em dog} and {\em house} a third way (see
Section~\ref{subsec:phrase-exper}). The composition is a construction that explicitly ties
together the two things that are being compared, and it depends on the nature of the comparison
that is desired, the task that is to be performed. We hypothesize that no single
stand-alone, task-independent representation can be constructed that is suitable for all purposes.

As we noted in the introduction, composition of vectors can result in a stand-alone
representation of a phrase, but composing similarities necessarily yields a linking
structure that connects a phrase to other phrases. These linking structures can
be seen in Figures \ref{fig:simr} to \ref{fig:3sims}. Intuitively, it seems that
an important part of how we understand a phrase is by connecting it to other phrases.
Part of our understanding of {\em dog house} is its connection to {\em kennel}. Dictionaries
make these kinds of connections explicit. From this perspective, the idea of an explicit
linking structure seems natural, given that making connnections among words and phrases is
an essential aspect of meaning and understanding.

\subsection{General Form of Similarities in the Dual-Space Model}
\label{subsec:formality}

In this subsection, we present a general scheme that ties together the
various similarities that were defined in Section~\ref{sec:experiments}.
This scheme includes similarities between chunks of text of arbitrary size.
The scheme encompasses phrasal similarity, relational similarity, and compositional
similarity.

Let $\mathbf{t}$ be a chunk of text (an ordered set of words),
$\langle t_{1}, t_{2}, \dots, t_{n}\rangle$, where each $t_i$ is a word.
We represent the semantics of $\mathbf{t}$ by $T = \langle \mathbf{D}, \mathbf{F} \rangle$,
where $\mathbf{D}$ and $\mathbf{F}$ are matrices. Each row vector
$\mathbf{d}_i$ in $\mathbf{D}$, $i = 1, 2, \dots, n$, is the row vector in domain
space that represents the domain semantics of the word $t_i$. Each row vector
$\mathbf{f}_i$ in $\mathbf{F}$, $i = 1, 2, \dots, n$, is the row vector in function
space that represents the function semantics of the word $t_i$. To keep the
notation simple, the parameters, $k_{\rm d}$ and $p_{\rm d}$ for domain space and
$k_{\rm f}$ and $p_{\rm f}$ for function space, are implicit.
Assume that the row vectors in $\mathbf{D}$ and $\mathbf{F}$ are normalized
to unit length. Note that the size of the representation $T$ scales linearly
with $n$, the number of words in $\mathbf{t}$, hence we have {\em information scalability}.
For large values of $n$, there will inevitably be duplicate words in $\mathbf{t}$, so
the representation could easily be compressed to sublinear size without loss
of information.

Let $\mathbf{t}_1$ and $\mathbf{t}_2$ be two chunks of text
with representations $T_1 = \langle \mathbf{D}_1, \mathbf{F}_1 \rangle$
and $T_2 = \langle \mathbf{D}_2, \mathbf{F}_2 \rangle$, where $\mathbf{t}_1$
contains $n_1$ words and $\mathbf{t}_2$ has $n_2$ words. Let $\mathbf{D}_1$
and $\mathbf{D}_2$ have the same parameters, $k_{\rm d}$ and $p_{\rm d}$, and let
$\mathbf{F}_1$ and $\mathbf{F}_2$ have the same parameters, $k_{\rm f}$ and $p_{\rm f}$.
Then $\mathbf{D}_1$ is $n_1 \times k_{\rm d}$, $\mathbf{D}_2$ is $n_2 \times k_{\rm d}$,
$\mathbf{F}_1$ is $n_1 \times k_{\rm f}$, and $\mathbf{F}_2$ is $n_2 \times k_{\rm f}$.
Note that $\mathbf{D}_1 \mathbf{D}_1^\mathsf{T}$ is an $n_1 \times n_1$ matrix
of the cosines between any two row vectors in $\mathbf{D}_1$. That is, the element
in the $i$-th row and $j$-th column of $\mathbf{D}_1 \mathbf{D}_1^\mathsf{T}$
is ${\rm cos}(\mathbf{d}_i, \mathbf{d}_j)$. Likewise, $\mathbf{D}_1 \mathbf{D}_2^\mathsf{T}$
is an $n_1 \times n_2$ matrix of the cosines between any row vector in
$\mathbf{D}_1$ and any row vector in $\mathbf{D}_2$.

Suppose that we wish to measure the similarity, ${\rm sim} ( \mathbf{t}_1 , \mathbf{t}_2)$,
between the two chunks of text, $\mathbf{t}_1$ and $\mathbf{t}_2$. In this paper, we have
restricted the similarity measures to the following general form:

\begin{equation}
\label{eqn:general}
{\rm sim} ( \mathbf{t}_1 , \mathbf{t}_2) = f(
\mathbf{D}_1 \mathbf{D}_1^\mathsf{T},
\mathbf{D}_1 \mathbf{D}_2^\mathsf{T},
\mathbf{D}_2 \mathbf{D}_2^\mathsf{T},
\mathbf{F}_1 \mathbf{F}_1^\mathsf{T},
\mathbf{F}_1 \mathbf{F}_2^\mathsf{T},
\mathbf{F}_2 \mathbf{F}_2^\mathsf{T} )
\end{equation}

\noindent In other words, the only input to the composition function $f$ is cosines
(and the implicit parameters, $k_{\rm d}$, $p_{\rm d}$, $k_{\rm f}$, and $p_{\rm f}$);
$f$ does not operate directly on any of the row vectors in $\mathbf{D}_1$, $\mathbf{D}_2$,
$\mathbf{F}_1$, and $\mathbf{F}_2$. In contrast to much of the work discussed in
Section~\ref{subsec:composition}, the composition operation is shifted out of the
representations, $T_1$ and $T_2$, and into the similarity measure, $f$. The exact
specification of $f$ depends on the task at hand. When $T_1$ and $T_2$ are sentences, we
envision that the structure of $f$ will be determined by the syntactic structures of
the two sentences.\footnote{Note that there is no requirement for the two
chunks of text, $\mathbf{t}_1$ and $\mathbf{t}_2$, to have the same number of words.
That is, $n_1$ does not necessarily equal $n_2$. In Section~\ref{subsec:wordnet-exper},
$n_1 \ne n_2$.}

Consider relational similarity (Section~\ref{subsec:sat-exper}):

\begin{align}
{\rm sim_1}(a\!:\!b, c\!:\!d) & = {\rm geo}({\rm sim_f}(a,c), {\rm sim_f}(b,d)) \\
{\rm sim_2}(a\!:\!b, c\!:\!d) & = {\rm geo}({\rm sim_d}(a,b), {\rm sim_d}(c,d)) \\
{\rm sim_3}(a\!:\!b, c\!:\!d) & = {\rm geo}({\rm sim_d}(a,d), {\rm sim_d}(c,b)) \\
{\rm sim_r}(a\!:\!b, c\!:\!d) & =
\left\{
\begin{array}{rl}
{\rm sim_1}(a\!:\!b, c\!:\!d) & \mbox{if ${\rm sim_2}(a\!:\!b, c\!:\!d) \ge {\rm sim_3}(a\!:\!b, c\!:\!d)$} \\
0 & \mbox{otherwise}
\end{array}
\right.
\end{align}

\noindent This fits the form of Equation~\ref{eqn:general} when we have
$\mathbf{t}_1 = \langle a, b \rangle$ and $\mathbf{t}_2 = \langle c, d \rangle$.
We can see that ${\rm sim_1}$ is based on cosines from $\mathbf{F}_1 \mathbf{F}_2^\mathsf{T}$,
${\rm sim_2}$ is based on cosines from $\mathbf{D}_1 \mathbf{D}_1^\mathsf{T}$ and
$\mathbf{D}_2 \mathbf{D}_2^\mathsf{T}$, and ${\rm sim_3}$ is based on cosines from
$\mathbf{D}_1 \mathbf{D}_2^\mathsf{T}$.

Consider compositional similarity (Section~\ref{subsec:wordnet-exper}):

\begin{align}
{\rm sim_1}(ab, c) & =
{\rm geo}({\rm sim_d}(a,c), {\rm sim_d}(b,c), {\rm sim_f}(b,c)) \\
{\rm sim_c}(ab, c) & =
\left\{
\begin{array}{rl}
{\rm sim_1}(ab, c) & \mbox{if $a \ne c$ and $b \ne c$} \\
0 & \mbox{otherwise}
\end{array}
\right.
\end{align}

\noindent This can be seen as an instance of Equation~\ref{eqn:general} in which
$\mathbf{t}_1 = \langle a, b \rangle$ and $\mathbf{t}_2 = \langle c \rangle$.
In this case,  ${\rm sim_1}$ is based on cosines from
$\mathbf{D}_1 \mathbf{D}_2^\mathsf{T}$ and $\mathbf{F}_1 \mathbf{F}_2^\mathsf{T}$.
The constraints, $a \ne c$ and $b \ne c$, can be expressed in terms
of cosines from $\mathbf{D}_1 \mathbf{D}_2^\mathsf{T}$, as
${\rm sim_d}(a,c) \ne 1$ and ${\rm sim_d}(b,c) \ne 1$. (Equivalently, we
could use cosines from $\mathbf{F}_1 \mathbf{F}_2^\mathsf{T}$.)
Similar analyses apply to the similarities in Sections \ref{subsec:phrase-exper}
and \ref{subsec:chiarello-exper}; these similarities are also instances
of Equation~\ref{eqn:general}.

Although the representations $T_1$ and $T_2$ have sizes that are linear functions
of the numbers of phrases in $\mathbf{t}_1$ and $\mathbf{t}_2$, the size of the composition
in Equation~\ref{eqn:general} is a quadratic function of the numbers of phrases in
$\mathbf{t}_1$ and $\mathbf{t}_2$. However, specific instances of this general
equation may be less than quadratic in size, and it may be possible to limit
the growth to a linear function. Also, in general, quadratic growth is often
acceptable in practical applications \cite{garey79}.

With function words (e.g., prepositions, conjunctions), one option would be
to treat them the same as any other words. They would be represented by vectors
and their similarities would be calculated in function and domain spaces.
Another possibility would be to use function words as hints to guide the
construction of the composition function $f$. The function words would not
correspond to vectors; instead they would contribute to determining the
linking structure that connects the two given chunks of text. The first option
appears more elegant, but the choice between the options should be made empirically.

\subsection{Automatic Composition of Similarities}
\label{subsec:auto}

In Section~\ref{sec:experiments}, we manually constructed the functions that
combined the similarity measures, using our intuition and background knowledge.
Manual construction will not scale up to the task of comparing any two
arbitrarily chosen sentences. However, there are good reasons for believing
that the construction of composition functions can be automated.

\citeA{turney08b} presents an algorithm for solving analogical mapping problems,
such as the analogy between the solar system and the Rutherford-Bohr model of the atom.
Given a list of terms from the solar system domain, \{{\em planet, attracts, revolves,
sun, gravity, solar system, mass}\}, and a list of terms from the atomic domain,
\{{\em revolves, atom, attracts, electromagnetism, nucleus, charge, electron}\},
it can automatically generate a one-to-one mapping from one domain to the other,
\{{\em solar system $\rightarrow$ atom, sun $\rightarrow$ nucleus, planet $\rightarrow$ electron,
mass $\rightarrow$ charge, attracts $\rightarrow$ attracts, revolves $\rightarrow$ revolves,
gravity $\rightarrow$ electromagnetism}\}. On twenty analogical mapping problems, it
attains an accuracy of 91.5\%, compared to an average human accuracy of 87.6\%.

The algorithm scores the quality of a candidate analogical mapping
by composing the similarities of the mapped terms. The composition function
is addition and the individual component similarities are holistic relational similarities.
The algorithm searches through the space of possible mappings for the mapping that
maximizes the composite similarity measure. That is, analogical mapping
is treated as an {\em argmax} problem, where the argument to be maximized
is a mapping function. In effect, the output of the algorithm (an analogical
mapping) is an automically generated composition of similarities.
The mapping structures found by the algorithm are essentially the same as the
linking structures that we see in Figures \ref{fig:simr} to \ref{fig:3sims}.

We believe that a variation of \citeS{turney08b} algorithm could be used to
automatically compose similarities in the dual-space model; for example, it should be
possible to identify paraphrases using automatic similarity composition. The proposal
is to search for a composition that maximizes composite similarity, subject
to various constraints (such as constraints based on the syntax of the
sentences). \citeA{turney08b} points out that analogical mapping could be used to align
the words in two sentences, but does not experimentally evaluate this suggestion.

Recent work \cite{lin11} has shown that {\em argmax} problems can be solved
efficiently and effectively if they can be framed as monotone submodular function
maximization problems. We believe that automatic composition of similarities
can fit naturally into this framework, which would result in highly scalable
algorithms for semantic composition.

Regarding information scalability, the dual-space model does not suffer from
information loss (unlike approaches that represent compositions with vectors
of fixed dimensionality), because the sizes of the representations grow
as the lengths of the phrases grow. The growth might be quadratic, but it
is not exponential. There are questions about how to automate composition
of similarities, which may have an impact on the computational complexity
of scaling to longer phrases, but there is evidence that these questions are tractable.

\section{Limitations and Future Work}
\label{sec:future}

One area for future work is to experiment with longer phrases (more than two
words) and sentences, as discussed in Section~\ref{subsec:auto}. An interesting
topic for research is how parsing might be used to constrain the automatic
search for similarity composition functions.

Here we have focused on two spaces, domain and function, but it seems likely
to us that a model with more spaces would yield better performance. We are currently
experimenting with a quad-space model that includes domain (noun-based contextual patterns),
function (verb-based), quality (adjective-based), and manner (adverb-based) spaces.
The preliminary results with quad-space are promising. Quad-space seems to be
related to \citeS{pustejovsky91} four-part qualia structure.

Another issue we have avoided here is morphology. As discussed in Section~\ref{subsec:using},
we used the {\em validForms} function in the WordNet::QueryData Perl interface to WordNet
to map morphological variations of words to their base forms. This implies that,
for example, a singular noun and its plural form should have the same semantic
representation. This is certainly a simplification and a more sophisticated model
would use different representations for different morphological forms of a word.

We have also avoided the issue of polysemy. It should be possible to extend past
work with polysemy in VSMs to the dual-space model \cite{schutze98,pantel02a,erk08}.

In this paper, we have treated the holistic model and the dual-space model
as if they are competitors, but there are certain cases, such as idiomatic
expressions, where the holistic approach is required. Likewise, the holistic
approach is limited by its inability to handle linguistic creativity.
These considerations suggest that the holistic and dual-space models must be
integrated. This is another topic for future work.

Arguably it is a limitation of the dual-space model that there are four
parameters to tune ($k_{\rm d}$, $p_{\rm d}$, $k_{\rm f}$, and $p_{\rm f}$).
On the other hand, perhaps any model with adaptive capacity must have
some parameters to tune. Further research is needed.

A number of design decisions were made in the construction of domain and function
space, especially in the conversion of phrases to contextual patterns (Sections
\ref{subsec:domain} and \ref{subsec:function}). These decisions were guided by our
intuitions. We expect that the exploration and experimental evaluation of this
design space will be a fruitful area for future research.

The construction of function space (Section~\ref{subsec:function}) is
specific to English. It may generalize readily to other Indo-European
languages, but some other languages may present a challenge. This is another
topic for future research.

Most of our composite similarities use the geometric mean to combine
domain and function similarities, but we see no reason to restrict the
possible composition functions. Equation~\ref{eqn:general} allows any
composition function $f$. Exploring the space of possible composition functions
is another topic for future work.

Another question is how formal logic and textual entailment can be integrated
into this approach. The dual-space model seems to be suitable for recognizing
paraphrases, but there is no obvious way to handle entailment. More generally,
we have focused on various kinds of similarity, but when we scale up from phrases
({\em red ball}) to sentences ({\em The ball is red}), we encounter truth and falsity.
\citeA{gardenfors04} argues that spatial models are a bridge between low-level
connectionist models and high-level symbolic models. He claims that spatial models
are best for questions about similarity and symbolic models are best for
questions about truth. We do not yet know how to join these two kinds of models.

\section{Conclusions}
\label{sec:conclusions}

The goal in this research has been to develop a model that unifies semantic
relations and compositions, while also addressing linguistic creativity, order
sensitivity, adaptive capacity, and information scalability. We believe that the
dual-space model achieves this goal, although there is certainly room for
improvement and further research.

There are many kinds of word--context matrices, based on various notions of context;
\citeA{sahlgren06} gives a good overview of the types of context that have been
explored in past work. The novelty of the dual-space model is that it includes
two distinct and complementary word--context matrices that work together synergistically.

With two distinct spaces, we have two distinct similarity measures, which can be
combined in many different ways. With multiple similarity measures, similarity
composition becomes a viable alternative to vector composition. For example, instead
of multiplying vectors, such as $\mathbf{c} = \mathbf{a} \odot \mathbf{b}$,
we can multiply similarities, such as
${\rm sim_{sa}}(a, b) = {\rm geo}({\rm sim_d}(a, b) , {\rm sim_f}(a, b))$.
The results here suggest that this is a fruitful new way to look at some of
the problems of semantics.

\acks{Thanks to George Foster, Yair Neuman, David Jurgens, and the reviewers of
{\em JAIR} for their very helpful comments on an earlier version of this paper.
Thanks to Charles Clarke for the corpus used to build the three spaces,
to Stefan B{\"u}ttcher for Wumpus, to the creators of WordNet for
making their lexicon available, to the developers of OpenNLP, to
Doug Rohde for SVDLIBC, to Jeff Mitchell and Mirella Lapata for sharing their
data and answering questions about their evaluation methodology, to Christine
Chiarello, Curt Burgess, Lorie Richards, and Alma Pollock for making their data
available, to Jason Rennie for the WordNet::QueryData Perl interface to WordNet,
and to the developers of Perl Data Language.}

\vskip 0.2in

\bibliography{turney12a}
\bibliographystyle{theapa}

\end{document}